\documentclass{article}

\usepackage{arxiv}

\usepackage{microtype}
\usepackage{graphicx}
\usepackage{subcaption}
\usepackage{booktabs}
\usepackage{hyperref}
\usepackage{amsmath}
\usepackage{amssymb}
\usepackage{mathtools}
\usepackage{amsthm}
\usepackage[capitalize,noabbrev]{cleveref}
\usepackage{natbib}
\usepackage{tikz}
\usepackage{tikz-cd}
\usepackage{pgfplots}
\usepackage{algorithm}
\usepackage{algpseudocode}
\usepackage{xcolor}
\usepackage{enumitem}
\RequirePackage{doi}
\usetikzlibrary{arrows.meta,calc,positioning,decorations.pathreplacing,fit}
\pgfplotsset{compat=1.18}

\algrenewcommand\algorithmicrequire{\textbf{Input:}}
\algrenewcommand\algorithmicensure{\textbf{Output:}}

\hypersetup{
  pdftitle={Prometheus: Automating Deep Causal Research Integrating Text, Data, and Scientific Models},
  pdfsubject={Causal research from language},
  pdfauthor={Anonymous Authors},
  pdfkeywords={causal extraction, predictive state representations, sheaves, topos world models, literature review},
  colorlinks=true,
  linkcolor=blue!55!black,
  citecolor=blue!55!black,
  urlcolor=blue!55!black
}

\theoremstyle{plain}
\newtheorem{theorem}{Theorem}[section]
\newtheorem{proposition}[theorem]{Proposition}

\theoremstyle{definition}
\newtheorem{definition}[theorem]{Definition}

\theoremstyle{remark}
\newtheorem{remark}[theorem]{Remark}

\newcommand{\C}{\mathcal{C}}

\newcommand{\Psh}{\mathcal{P}}

\newcommand{\doop}{\operatorname{do}}

\title{\textsc{Prometheus}:  Automating Deep Causal Research Integrating Text, Data, and Scientific Models\thanks{Draft under revision.}}

\author{ Sridhar Mahadevan \\
	Adobe Research and University of Massachusetts, Amherst\\
	\texttt{smahadev@adobe.com, mahadeva@umass.edu}
}

\date{}

\begin{document}
\maketitle

\begin{abstract}
Large language models can extract local causal claims from text, but those claims
become more useful when organized as persistent, navigable world models rather
than as flat summaries.  We introduce \textsc{Prometheus}, a framework that turns
retrieved literature, filings, reviews, reports, agent traces, source data,
code, simulations, and scientific models into \emph{causal atlases}:
sheaf-like families of local causal predictive-state models over an explicit
cover of a research substrate.  Each local region contains causal
episodes, structured claim tables, predictive tests, support statistics, and
provenance; restriction maps compare overlapping regions; gluing diagnostics
expose agreement, drift, contradiction, and underdetermination.  The resulting
Topos World Model is not a single universal graph.  It is a research instrument
for navigating what a corpus says, where it says it, how strongly it is
supported, and where local claims fail to assemble into a coherent global view.
We describe the \textsc{Prometheus} pipeline, formalize causal episodes and local
predictive-state sheaves, present the Claims Atlas interface, and propose an
evaluation program centered on coverage, drift visibility, provenance,
support aggregation, expert navigation time, and rerun consistency.  Three
literature-atlas case studies---ocean-temperature impacts on marine
populations, GLP-1 weight-loss evidence, and resveratrol/red-wine health-benefit
claims---illustrate deep causal research from text with explicit locality,
evidence, persistent state, and gluing tension.  Four grounded-counterfactual
case studies---a Nature Climate Change microplastics forcing paper, an Indus
Valley hydrology paper with VIC-derived figure data and model code, and the
canonical Sachs protein-signaling study with single-cell perturbation data, and
a Nature singing-mouse study with MAPseq projection matrices---show a stronger
mode: when a paper ships source data, simulation
outputs, or code, \textsc{Prometheus} can evaluate a counterfactual against
that scientific substrate and then rebuild the sheaf world model around the
measured intervention result.
\end{abstract}


\section{Introduction}

Large language models are now capable summarizers of scientific papers,
financial filings, product reviews, and operational records.  They can retrieve
relevant passages, restate conclusions, and extract many local causal claims.
For deep research, however, this is not enough.  A researcher often needs to
know which causal claims recur across a corpus, which claims are regime-specific,
which apparent disagreements are due to different populations or measurements,
and where a literature has enough support to justify a follow-up query,
experiment, or decision.  Ordinary summaries are too flat for this task.

\textsc{Prometheus} starts from the thesis that language-derived causal knowledge should
be represented as a family of local predictive models over a corpus.  A paper,
section, time window, population, product-use context, or regulatory workflow
can each induce a local model.  These local models overlap.  When their shared
predictions and causal claims agree, they may glue into larger coherent regions.
When they do not, the disagreement is not noise to be averaged away; it is a
research signal.  It may indicate drift, confounding, incompatible measurement,
source-quality variation, or a genuine regime boundary.

We call the resulting object a \emph{Topos World Model}.  Operationally, it is a
sheaf-like causal atlas: local predictive-state representations indexed by
contexts, connected by restriction maps, and annotated with support, gluing
tension, drift, and provenance.  The atlas is designed for use by a human or
agentic researcher who wants to ask: What is the main causal spine of this
literature?  Which regions support it?  Which contexts break it?  Which passages
and tables justify a local claim?  What changes between two retrieval runs or
two time periods?

This paper positions \textsc{Prometheus} as a research instrument rather than as a claim
that a single benchmark score captures the value of the system.  The intended
contribution is causal extraction plus topological organization plus navigable
evidence.  The resulting object is not merely a knowledge graph, not merely a
retrieval-augmented summary, and not merely a structural causal model.  It is a
finite, inspectable approximation to a sheaf of local causal predictive states
over text.

\paragraph{Contributions.}
This paper makes six contributions.
\begin{enumerate}[leftmargin=*]
  \item We introduce \textsc{Prometheus}, a language-to-Topos-World-Model pipeline for
  deep causal research from text.
  \item We extend the earlier \textsc{Democritus} line of causal claim extraction and
  cSQL-style causal tables into local causal predictive-state models with
  covers, restrictions, gluing diagnostics, persistent states, and causal
  atlases.
  \item We formalize causal events, episodes, contexts, covers, restrictions,
  local predictive-state tables, and gluing tensions in a finite sheaf-theoretic
  setting suitable for implementation.
  \item We describe the Claims Atlas, an interface object that organizes a
  corpus by causal spine, local regions, support, drift, regime tension, and
  provenance drill-downs.
  \item We propose an evaluation program appropriate to causal research
  instruments: not only extraction accuracy, but also coverage, drift
  visibility, support aggregation, provenance quality, expert navigation time,
  and consistency across reruns.
  \item We demonstrate grounded counterfactual layers in four scientific
  domains: a Nature Climate Change microplastics paper whose source tables
  support an optical-forcing intervention, and an Indus Valley hydrology paper
  whose VIC-derived figure data and model code support a drought-restoration
  intervention, and the Sachs protein-signaling benchmark whose single-cell
  environment panel supports an experimental-regime substitution, and a Nature
  singing-mouse study whose MAPseq projection matrices support a species-level
  projection-attenuation intervention.  In each case, modified causal
  observations are used to rebuild the sheaf world model.
\end{enumerate}

\section{Related Work}
\label{sec:related}

Retrieval-augmented generation improves factual grounding by conditioning
answers on retrieved passages \citep{lewis2020retrieval}.  Causal extraction
pipelines go further by identifying cause-effect statements in text
\citep{girju2003automatic,hendrickx2010semeval}.  Yet both approaches often collapse
the corpus into an answer.  This is precisely what a deep researcher cannot
afford to lose.

Consider a corpus on ocean warming and fish populations.  One region of the
literature may emphasize thermal stress and migration.  Another may focus on
food-web disruption.  A third may show adaptation or local resilience in
particular species.  A flat answer such as ``warming reduces fish populations''
is directionally useful but structurally poor.  It hides the population,
location, temperature range, time scale, measurement protocol, and ecological
mediators that decide whether the claim should be transported to a new case.

The same issue appears in product reviews.  A shoe can be comfortable for short
runs but painful for long mileage; waterproof in light rain but poor after
repeated washing; well-rated overall but return-prone in a narrow sizing
regime.  In SEC or operational workflows, a filing may describe investment,
optimization, supply-chain risk, regulatory exposure, and expected margin
effects, but each relationship may only be valid under specific market and
time-window assumptions.

\textsc{Prometheus} therefore treats each text-derived claim as local.  It asks where the
claim lives, which neighboring contexts it overlaps, which tests it predicts,
and whether its restrictions agree with nearby local models.

\paragraph{Causal relation extraction from text.}
There is a long line of work on identifying causal relations in natural
language, from cue-phrase and pattern-based systems to neural classifiers; see
surveys of causal relation extraction and event causality identification
\citep{yang2022causalSurvey,he2022eventCausalitySurvey}.  Classical systems
usually predict whether a pair of spans or events in a sentence stands in a
causal relation, while corpus-scale work connects such local predictions to
event forecasting or explanatory retrieval \citep{radinsky2012learningCausality}.
\textsc{Prometheus} uses such extracted relations as evidence units, but the
paper's object of study is downstream: how thousands of local claims should be
localized, compared, transported, or blocked across corpus regions.

\paragraph{Causal knowledge bases and graphs from corpora.}
Causal knowledge-base projects mine cause--effect tuples from large corpora and
aggregate them into graph-structured resources
\citep{hassanzadeh2020causalKB}.  This graph-building perspective is close to
the first \textsc{Democritus} contribution, where LLM-generated causal
statements are compiled into local causal models and larger causal atlases
\citep{mahadevan2025largecausalmodelslarge}.  \textsc{Prometheus} extends that line by
treating local graphs and cSQL rows as observations for local causal PSRs.  The
global object is therefore not one merged graph but a sheaf-like family of
charts whose overlaps reveal agreement, drift, contradiction, and
underdetermination.

\paragraph{LLMs for causal discovery and reasoning.}
A growing literature asks whether LLMs can propose causal directions, graph
structures, interventions, or counterfactual explanations from variable
descriptions and textual context
\citep{kiciman2024causalLLM,le2024multiagentCausalDiscovery}.
\textsc{Prometheus} is deliberately more conservative.  It does not treat the
LLM as an oracle for ground-truth causal discovery.  Instead, the LLM helps
surface causal discourse: claims, mechanisms, modifiers, regimes, and source
passages that can be normalized, audited, and compared.  Local intervention
probes in the atlas are therefore model-internal research tests unless paired
with external data and identification assumptions.

\paragraph{Agentic systems for automated scientific discovery.}
Recent systems also aim to automate larger portions of the scientific workflow.
The AI Scientist-v2, for example, uses agentic tree search to propose
hypotheses, design and execute machine-learning experiments, analyze and
visualize results, and write scientific manuscripts
\citep{yamada2025aiScientistV2}.  This line of work is close in ambition to
\textsc{Prometheus}: both ask how AI systems can participate in scientific
discovery rather than merely answer questions about existing papers.  The
emphasis is different.  AI Scientist-v2 organizes autonomous experimentation
and manuscript generation, primarily in machine-learning research settings.
\textsc{Prometheus} instead constructs an explicit causal topos world model
from heterogeneous research artifacts---text, data, figures, source code, and
scientific models---so that local claims, gluing failures, evidentiary limits,
and grounded counterfactual revisions remain inspectable.  In this sense,
\textsc{Prometheus} can be viewed as a complementary world-model layer for
scientific agents: it records what a research substrate supports, where it
does not glue, and which counterfactuals can actually be evaluated.

\paragraph{Causality-aware NLP.}
More broadly, causal ideas have been used to study text effects,
counterfactual augmentation, representation robustness, and explanations for
NLP systems \citep{jin2021causalnlpSurvey}.  \textsc{Prometheus} points in the
opposite direction: it uses NLP and LLM extraction to construct explicit causal
artifacts for human research.  The Claims Atlas is meant to be inspected,
corrected, extended, and rerun, so provenance and gluing failures are part of
the output rather than post-hoc debugging aids.

\section{From \textsc{Democritus} to \textsc{Prometheus}}

\textsc{Democritus} is the predecessor pipeline: a language-to-causal-model
system for compiling documents into local causal models, causal databases, and
interactive diagnostic artifacts
\citep{mahadevan2025largecausalmodelslarge}.  A public implementation of the
released \textsc{Democritus} client is available as
\texttt{Democritus\_OpenAI} \citep{mahadevanDemocritusOpenAI}.  The broader categorical machine
learning background for this line of work is developed in
\citet{mahadevanCatAGIBook}.  It extracts
local causal claims, organizes them into causal triples or local causal models,
and stores structured outputs in cSQL-like causal tables.  This is already
useful: it turns unstructured text into queryable causal objects.

\textsc{Prometheus} changes the central object.  Instead of treating a local DAG or
causal table as the final representation, \textsc{Prometheus} treats it as evidence for a
local predictive-state model.  A local model records not only that \(X\) causes
or influences \(Y\), but which histories and tests are present, what the model
predicts under those tests, how much support each cell has, and where the
evidence came from.  The global object is not one merged DAG.  It is a sheaf-like
family of local models, equipped with restriction and gluing diagnostics.

\begin{table}[t]
\centering
\begin{tabular}{lll}
\toprule
System & Local object & Global object \\
\midrule
\textsc{Democritus} & Causal claim, DAG, cSQL row & Causal synthesis over extracted tables \\
\textsc{Prometheus} & Causal PSR over a context & Topos World Model / causal atlas \\
\bottomrule
\end{tabular}
\caption{\textsc{Prometheus} inherits the extraction discipline of \textsc{Democritus} but shifts
the representation from graph-centered synthesis to predictive-state sheaves.}
\label{tab:democritus-prometheus}
\end{table}

This shift matters because deep research is rarely about finding one graph.  It
is about understanding which local graphs, claims, and predictions are
transportable.  Topological organization gives the system a way to say: these
regions agree, these overlap but pull apart, and this claim should not be moved
without additional evidence.

\section{Which Parts of the \textsc{Democritus} Pipeline Are Reused?}
\label{sec:democritus-pipeline-map}

The original \textsc{Democritus} arXiv paper described a six-module pipeline for
constructing large causal models from language
\citep{mahadevan2025largecausalmodelslarge}.  \textsc{Prometheus} should be read as a
continuation of that pipeline, but not as a simple rebranding of it.  The first
four modules still provide the extraction discipline: they turn a seed research
domain or corpus into topics, causal questions, causal statements, and typed
relational triples.  The new contribution begins when those triples and cSQL
rows are no longer treated as the terminal graph artifact.  Instead, they become
observations from which \textsc{Prometheus} constructs local causal predictive-state
models indexed by an explicit cover.

\begin{table}[t]
\centering
\small
\begin{tabular}{p{0.22\linewidth}p{0.35\linewidth}p{0.33\linewidth}}
\toprule
\textsc{Democritus} module & Original role & Role in \textsc{Prometheus} \\
\midrule
Module 1: Topic graph &
LLM-driven breadth-first expansion of a domain into a topic hierarchy. &
Optional for open-ended research runs; replaced or supplemented by retrieval
clusters when a concrete paper/review/filing corpus is already supplied. \\
Module 2: Causal questions &
Generate causal questions for each topic. &
Used to seed query expansion, retrieval, and local test templates; questions
also become human-readable atlas entry points. \\
Module 3: Causal statements &
Generate or extract causal statements and explanations. &
Used directly, but grounded more heavily in acquired evidence units and
provenance rather than free generation alone. \\
Module 4: Relational triples &
Extract subject--relation--object triples and build a multi-relational causal
graph. &
Retained as the cSQL/claim layer: cause, effect, mediator, modifier, polarity,
context, support, and provenance. \\
Module 5: Relational manifold &
Apply a Geometric Transformer and UMAP to organize the large causal graph. &
Used selectively as a geometric diagnostic or viewer; the central state becomes
local PSR tables rather than only a node embedding manifold. \\
Module 6: Topos slice and unification &
Store domain slices and prepare them for cross-slice topos reasoning. &
Replaced by a Topos World Model artifact: local causal PSRs, restriction maps,
gluing diagnostics, persistent state, and Claims Atlas navigation. \\
\bottomrule
\end{tabular}
\caption{How \textsc{Prometheus} reuses and extends the six-module \textsc{Democritus}
pipeline.  Modules 1--4 supply causal extraction and normalization; the new
Topos World Model construction begins when extracted claims are assembled into
local predictive states over a cover.}
\label{tab:democritus-six-stage-map}
\end{table}

This mapping clarifies the boundary between extraction and world modeling.  In
\textsc{Democritus}, Module~4 produces a relational graph, Module~5 geometrically
organizes that graph, and Module~6 stores the result as a topos slice.  In
\textsc{Prometheus}, the graph is an intermediate observation layer.  A triple such as
\[
  (\text{food limitation},\ \texttt{reduces},\ \text{thermal tolerance})
\]
is not merely an edge in a global graph.  It is assigned to one or more local
contexts, connected to evidence units, folded into causal episodes, and used to
populate predictive tests such as
\[
  \text{food limitation}\to\text{thermal stress}
  \Rightarrow \text{larval-survival change}.
\]
The output is therefore a table-valued local section with support and
provenance, not only a point or edge in an embedding.

\paragraph{Stage A: inherited causal extraction.}
The inherited \textsc{Democritus} side of \textsc{Prometheus} performs topic or retrieval
expansion, causal-question generation, causal-statement extraction, and typed
triple normalization.  In open-ended exploratory runs, topic expansion still
acts as a breadth-first search over a domain.  In artifact-backed runs such as
the ocean-temperature case study, the retrieval layer supplies the primary
corpus and the Democritus-style modules operate on acquired documents,
paragraphs, tables, abstracts, and metadata.  Either way, this stage produces
the familiar claim substrate: causal rows, local graphs, relation types,
modifiers, confidence scores, and provenance links.

\paragraph{Stage B: context and episode construction.}
The first \textsc{Prometheus}-specific step is to stop treating all extracted triples as
commensurable.  Each claim is assigned to one or more contexts: document,
subtopic, species group, measurement regime, product-use stage, company
workflow, fiscal year, or agent role.  Claims are then assembled into causal
episodes
\[
  h=(e_1,\ldots,e_k),
\]
where each event \(e_i\) carries an action or condition, an observation or
outcome, time or ordering information when available, and provenance.  This is
the point at which a causal graph becomes a history/test object suitable for
predictive-state estimation.

\paragraph{Stage C: local predictive-state induction.}
For each context \(U\), \textsc{Prometheus} enumerates supported tests
\(\tau\), estimates \(M_U[h,\tau]\), and stores support and uncertainty.  The
claim table is still present, but it is now embedded inside a local causal PSR.
The practical effect is that a researcher can ask not only whether a claim was
extracted, but which future continuations it supports, how strongly it is
supported in a local regime, and whether neighboring regimes agree.

\paragraph{Stage D: topos world-model construction.}
The new topos layer begins with a cover of contexts and a family of local PSR
sections.  Restriction maps align shared histories, tests, claims, and
provenance across overlaps.  Gluing diagnostics then determine whether local
sections assemble into a larger section or whether their disagreement should be
recorded as contradiction, drift, regime dependence, or underdetermination.
Persistent state makes this object cumulative: later runs can add evidence,
compare against earlier sections, and report whether a tension was repaired or
made sharper.

Thus, \textsc{Prometheus} is best understood as:
\begin{align*}
  \textsc{Prometheus}
  &=
  \text{Democritus extraction}
  + \text{context-indexed causal PSRs} \\
  &\quad
  + \text{restriction/gluing diagnostics}
  + \text{persistent causal atlas}.
\end{align*}
The inherited modules supply breadth and causal normalization.  The new modules
supply locality, predictive state, transport tests, and explicit certificates of
non-gluing.

\section{The \textsc{Prometheus} Pipeline}

\begin{figure}[t]
\centering
\footnotesize
\begin{tikzpicture}[
  node distance=6mm,
  box/.style={draw,rounded corners,align=center,text width=0.86\linewidth,inner sep=5pt},
  arrow/.style={-Latex,thick}
]
\node[box] (q) {Research question and retrieved corpus};
\node[box,below=of q] (e) {Document acquisition: papers, filings, reviews, tables, reports};
\node[box,below=of e] (c) {Claim extraction: causal events, episodes, cSQL rows, provenance};
\node[box,below=of c] (cover) {Context cover: topics, documents, populations, time windows, workflows, regimes};
\node[box,below=of cover] (psr) {Local causal PSRs: histories, tests, predictions, support, uncertainty};
\node[box,below=of psr] (glue) {Restrictions and gluing diagnostics: agreement, drift, contradiction, underdetermination};
\node[box,below=of glue] (atlas) {Claims Atlas: causal spine, local regions, tensions, provenance drill-downs};
\draw[arrow] (q) -- (e);
\draw[arrow] (e) -- (c);
\draw[arrow] (c) -- (cover);
\draw[arrow] (cover) -- (psr);
\draw[arrow] (psr) -- (glue);
\draw[arrow] (glue) -- (atlas);
\end{tikzpicture}
\caption{\textsc{Prometheus} turns a corpus into a navigable causal atlas.}
\label{fig:pipeline}
\end{figure}
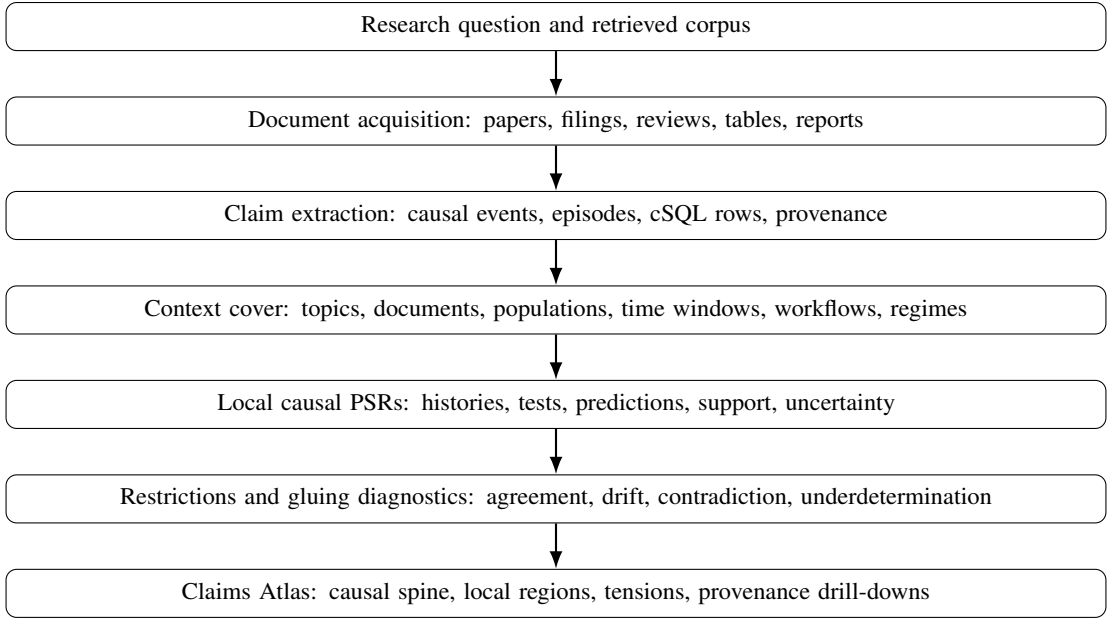

Figure~\ref{fig:pipeline} shows the pipeline.  \textsc{Prometheus} begins with a research
question or seed corpus.  A retrieval and acquisition layer gathers documents,
sections, passages, tables, and metadata.  Extraction converts these units into
causal events, episodes, claims, and cSQL-style rows with provenance.  A cover
constructor builds local contexts such as topic regions, document clusters,
time windows, populations, product-use situations, regulatory stages, or agent
roles.  Each context receives a local causal predictive-state representation.
Restriction maps compare overlaps, gluing diagnostics expose tensions, and the
Claims Atlas renders the resulting world model for research navigation.

\paragraph{Evidence units.}
An evidence unit may be a paper, paragraph, table row, review, filing section,
transcript segment, benchmark report, or agent trace.  Each extracted object
retains source identity, offsets or page references when available, retrieval
metadata, model prompts, and extraction confidence.

\paragraph{LLM backend and usage accounting.}
The current artifact runs use an OpenAI API-based LLM backend for extraction,
normalization, and local synthesis calls.  \textsc{Prometheus} records request
counts, token counts, and estimated API cost as run metadata, so the resulting
world model can be audited not only for evidence and provenance but also for
computational budget.  These numbers are implementation-dependent rather than
theoretical properties of the framework, but they are useful for reproducing
the scale of a run and comparing alternative extraction strategies.

\paragraph{Causal episodes.}
\textsc{Prometheus} represents text-derived causal structure as episodes:
\[
  h = (e_1,\ldots,e_k), \qquad
  e_i = (\text{actor}, \text{action/condition}, \text{observation}, t_i, p_i),
\]
where \(p_i\) is provenance.  Episodes can encode observational relations,
declared interventions, temporal order, and qualitative outcomes.  They are
converted into histories and tests for local predictive-state estimation.

\paragraph{cSQL and predictive state.}
cSQL tables store normalized causal rows: cause, effect, mediator, modifier,
polarity, strength, context, and provenance.  \textsc{Prometheus} keeps these rows, but
uses them to populate predictive-state tables.  A row such as ``thermal stress
reduces juvenile survival in species \(s\) under region \(r\)'' becomes a claim,
a context assignment, and a set of tests concerning survival, migration, and
population change under nearby histories.

\paragraph{Persistent state.}
A \textsc{Prometheus} run emits a durable world-model artifact.  Follow-up runs can be
conditioned on a previous state and compared against it.  This allows the system
to report whether new evidence stabilized a region, introduced drift, repaired a
gluing tension, or opened a new local context.

\section{From Text to Local Causal PSRs}
\label{sec:text-to-psr}

We now describe the predictive-state construction used by \textsc{Prometheus}.
Classical predictive state representations model a controlled process through
observable predictions rather than latent variables \citep{littman2001predictive,singh2004predictive}.  A
controlled history is a finite sequence
\[
  h=(a_1,o_1)\cdots(a_t,o_t),
\]
with actions \(a_i\in\mathcal{A}\) and observations \(o_i\in\mathcal{O}\).  A
future test \(\tau\) is an action--observation continuation, and the state at
\(h\) is the vector of probabilities \(\Pr(\tau\mid h)\) over a selected family
of tests.  Spectral PSR learning organizes these quantities into controlled
Hankel matrices
\[
  H_{p,\tau}=P(p\tau),
\]
whose rows are history prefixes and whose columns are suffix tests.  Under a
finite-rank assumption, a truncated factorization of \(H\) yields observable
operators \(W_{a,o}\) and a finite predictive realization.

\textsc{Prometheus} keeps this observable-prediction viewpoint but modifies the learning
recipe for text.  Literature and filing corpora do not give one clean controlled
trajectory with complete action--observation pairs.  They give many short,
heterogeneous fragments: claims, mechanisms, plans, measurements, caveats, and
outcomes.  A direct Hankel/SVD pipeline would therefore be brittle.  The
implemented estimator instead constructs a compressed, context-indexed Hankel
family from language-derived episodes.

\begin{algorithm}[t]
\caption{Language-to-local-PSR construction}
\label{alg:text-psr}
\begin{algorithmic}[1]
\Require Retrieved corpus \(D\), extracted evidence units \(E\), context cover \(\mathcal{U}\)
\Ensure Local PSRs \(\{\Psh(U)\}_{U\in\mathcal{U}}\), restrictions, gluing diagnostics
\State Extract causal events and normalized cSQL rows with provenance.
\State Assemble events into episodes \(h=(e_1,\ldots,e_k)\).
\State Assign each event and episode to one or more local contexts \(U\).
\State Enumerate supported compressed tests \(\tau\), e.g. \(a_1\to\cdots\to a_k\Rightarrow o\), cause--effect continuations, or mechanism--outcome motifs.
\State Estimate smoothed local probabilities \(\hat p_U(\tau)\) from support in \(U\), neighboring contexts, and the full corpus.
\State Form local PSR tables \(M_U[h,\tau]\) together with support, uncertainty, and provenance.
\State Build restriction maps by aligning shared histories, tests, claims, and provenance across overlaps.
\State Record gluing diagnostics from mismatch on shared cells or projected local sections.
\end{algorithmic}
\end{algorithm}

In the ocean-temperature artifact, tests are not product-review sentiment tests;
they are causal and mechanistic continuations such as thermal stress leading to
larval survival changes, food limitation reducing thermal tolerance, subpolar
gyre weakening affecting heat-wave regimes, or sea-grass habitat suitability
shifting under warming.  A local context \(U\) stores a table
\[
  M_U[h,\tau]\approx \hat p_U(\tau\mid h),
\]
plus support counts and provenance links.  When evidence is too sparse for a
literal conditional probability, \(M_U\) should be read as a normalized
predictive support score.  This is why the HTML artifact calls the result a
\emph{finite local predictive-state family}: it is an inspectable,
language-adapted PSR object, not an opaque embedding.

Restrictions and gluing are computed directly on overlapping test signatures.
If two contexts share histories or tests, \textsc{Prometheus} compares the corresponding
cells and reports mean and maximum gaps.  Compatible overlaps can be summarized
as larger sections; incompatible overlaps become obstruction data.  This is the
practical compromise made throughout this implementation: exact spectral operator
recovery is deferred, while the current system exposes the objects needed for
prediction, intervention-style comparison, and multi-context consistency.

\section{Formal Model}

We give a finite operational formalization, drawing on sheaf and topos semantics
\citep{maclane1992sheaves,abramsky2011sheaf} and the categorical machine
learning perspective developed in \citet{mahadevanCatAGIBook}.  The goal is not to claim that the
implementation realizes all of topos theory, but to make the modeling contract
precise enough to support inspection, testing, and extension.

\paragraph{Predictive contexts and covers.}
The basic modeling choice is to treat world modeling from text as a local
problem.  A context is not merely a document identifier; it is a region in which
the extracted claims are expected to have comparable semantics.  In the three
case studies, contexts include literature subtopics and ecological regimes,
filing workflow stages, product-use stages, and agent roles.  Covers encode the
comparability relation used for aggregation: two documents may cover a
literature theme, four workflow slices may cover a company-year, and a set of
sector-matched company-years may cover a fiscal-year regime.

\begin{definition}[Context site]
Let \(\C\) be a finite category of corpus contexts.  Objects \(U \in \C\) are
local regions such as documents, topics, time windows, populations, product-use
situations, workflow stages, or agent roles.  A morphism \(V \to U\) represents
inclusion, overlap, refinement, projection, or a declared translation.  A cover
\(\{U_i \to U\}_{i \in I}\) declares that the family \(U_i\) covers \(U\) for a
specified modeling purpose.
\end{definition}

In the ideal set-valued case, a model over \(\C\) is an object of the presheaf
topos \([\C^{op},\mathbf{Set}]\).  In the numerical implementation, the values
are finite tables, vectors, and metadata records, so it is often more accurate
to view the object as a presheaf into a finite data category or a vector-valued
functor category.  The set-valued topos supplies the semantics of locality and
gluing; the finite tables supply the artifact that users can inspect.

\begin{definition}[Local causal predictive-state representation]
For each context \(U\), \textsc{Prometheus} assigns a local object
\[
  \Psh(U) = (H_U,T_U,M_U,S_U,\Pi_U,D_U).
\]
Here \(H_U\) is a finite set of histories, \(T_U\) is a finite set of tests,
\(M_U[h,t]\) is a prediction or score for test \(t\) after history \(h\),
\(S_U[h,t]\) records support, \(\Pi_U[h,t]\) records provenance, and \(D_U\)
stores diagnostics such as sparsity, uncertainty, extraction confidence, and
local rank.
\end{definition}

This definition is inspired by predictive state representations
\citep{littman2001predictive,singh2004predictive}.  The difference is that
\textsc{Prometheus} learns or estimates many local PSRs from text and then studies their
compatibility over a context cover.

\paragraph{Connection to classical PSR learning.}
Classical spectral PSR learning starts with controlled histories
\[
  h=(a_1,o_1)\cdots(a_t,o_t)
\]
and future tests \(\tau\) consisting of action--observation continuations.  The
central empirical object is a controlled Hankel matrix
\[
  H_{p,\tau}=P(p\tau),
\]
where \(p\) ranges over prefixes and \(\tau\) over suffix tests.  Under a
finite-rank assumption, a truncated factorization of \(H\) yields a finite
predictive realization with observable operators \(W_{a,o}\).  \textsc{Prometheus}
uses this as a reference model rather than as a literal estimator.  Text corpora
rarely provide one synchronized controlled trajectory with complete
action--observation pairs; they provide heterogeneous fragments with missing
intermediate observations, uneven support, and context-dependent meanings.

The implemented estimator therefore replaces one global Hankel matrix by a
family of compressed local tables.  For a context \(U\), a compressed test has
the schematic form
\[
  \tau = a_1 \to \cdots \to a_k \Rightarrow o,
\]
where \(a_i\) may be a normalized action, condition, mechanism, or causal motif,
and \(o\) is a downstream observation or outcome label.  In the ocean case,
tests include mechanism--outcome continuations such as warming
\(\to\) thermal stress \(\Rightarrow\) larval-survival change; in filing and
review cases they correspond to workflow and usage continuations.

\paragraph{Local estimation.}
Let \(n_U(h,\tau)\) denote the support for test \(\tau\) after history \(h\) in
context \(U\), and let \(N_U(h)\) be the supported mass for histories comparable
to \(h\).  The simplest local estimate is a smoothed frequency
\[
  \hat p_U(\tau\mid h)
  =
  \frac{n_U(h,\tau)+\alpha \hat p_0(\tau)}
       {N_U(h)+\alpha},
\]
where \(\hat p_0\) is a corpus-, cover-, or domain-level backoff distribution.
The concrete pipelines use the same principle with several backoff levels.  For
example, in domains with nested context levels, the estimator can blend local,
neighboring-cover, and corpus-level support:
\[
  \hat p_U(\tau)
  =
  w_{\mathrm{loc}}(U,\tau)\hat p_{\mathrm{loc}}(\tau)
  +w_{\mathrm{nbr}}(U,\tau)\hat p_{\mathrm{nbr}}(\tau)
  +w_{\mathrm{corp}}(U,\tau)\hat p_{\mathrm{corp}}(\tau),
\]
with weights determined by available support.  The ocean-temperature artifact
uses analogous local, neighboring-context, and corpus backoff.  Each table cell
therefore carries both a predictive value and the evidential basis needed to
audit it: counts, smoothing/backoff level, extraction confidence, and
provenance.

\begin{definition}[Restriction map]
For a morphism \(r:V \to U\), a restriction map
\[
  \rho_{UV}: \Psh(U) \to \Psh(V)
\]
aligns histories, tests, claims, support, and provenance from \(U\) to \(V\).
In the finite implementation, \(\rho_{UV}\) may be a partial alignment map plus
a comparison of shared cells.
\end{definition}

Operationally, restriction has three parts.  First, histories and tests are
projected onto the vocabulary shared by the two contexts.  Second, claim and
provenance identifiers are aligned so that apparent agreement can be traced back
to evidence rather than only to surface labels.  Third, the resulting shared
cells are scored for mismatch.  If
\[
  \Omega_{UV}
  =
  (H_U\times T_U)\cap(H_V\times T_V)
\]
is the shared signature, a typical overlap discrepancy is
\[
  \Delta(U,V)
  =
  \frac{1}{|\Omega_{UV}|}
  \sum_{(h,\tau)\in\Omega_{UV}}
  \lambda_{h,\tau}
  \left|
    M_U[h,\tau]-M_V[h,\tau]
  \right|,
\]
where \(\lambda_{h,\tau}\) downweights unsupported or low-confidence cells.  The
artifact also records maxima and provenance-level explanations for high-gap
cells, because a small mean can hide a scientifically important contradiction.

\begin{definition}[Gluing tension]
For local sections \(s_i \in \Psh(U_i)\) over a cover \(\{U_i \to U\}\), the
pairwise gluing tension on overlaps is
\[
  \tau_{ij}
  =
  w_{ij}
  \left\|
    \rho_{U_i,U_i\cap U_j}(s_i)
    -
    \rho_{U_j,U_i\cap U_j}(s_j)
  \right\|^2,
\]
where \(w_{ij}\) is an overlap-confidence or support weight.  The total gluing
tension is \(\tau(\{s_i\})=\sum_{i<j}\tau_{ij}\).
\end{definition}

Low tension indicates that local models agree on shared tests or claims.  High
tension is classified by the atlas as contradiction, drift, regime dependence,
or underdetermination depending on support, directionality, time, and context
metadata.

\paragraph{Approximate sheaf condition.}
Exact sheaf gluing would require compatible local sections to determine a
unique global section.  \textsc{Prometheus} uses a finite tolerance version:
local sections are considered \(\epsilon\)-compatible on a cover when their
weighted overlap gaps are below a declared tolerance on sufficiently supported
shared signatures.  A candidate glued section is then formed only from the
compatible cells, for example by support-weighted aggregation,
\[
  M_U[h,\tau]
  =
  \frac{\sum_i \omega_i(h,\tau)M_{U_i}[h,\tau]}
       {\sum_i \omega_i(h,\tau)},
\]
where \(\omega_i\) combines support, extraction confidence, and relevance to the
cover.  Unsupported cells remain local.  Incompatible cells are not averaged
away; they become obstruction records with source passages and context
metadata.

\begin{proposition}[Operational sheaf condition]
If all local sections over a declared cover have compatible restrictions on
overlaps and the support of the shared cells exceeds a user-specified threshold,
then \textup{\textsc{Prometheus}} may construct a glued section over \(U\).  If compatibility
fails, \textup{\textsc{Prometheus}} does not force a global merge; it records the obstruction as a
gluing diagnostic.
\end{proposition}

\begin{remark}
This operational condition is intentionally conservative.  In research use, a
failed gluing attempt is often more valuable than a brittle averaged answer.
\end{remark}

\paragraph{Two-stage gluing.}
The same formalism can be iterated across levels of analysis.  In the filing
experiments, the local workflow slices
\[
  x_{C,y}^{\mathrm{ops}},\quad
  x_{C,y}^{\mathrm{mkt}},\quad
  x_{C,y}^{\mathrm{fin}},\quad
  x_{C,y}^{\mathrm{inn}}
\]
may first glue into a company-year section \(s_{C,y}\).  These company-year
sections can then be compared over a second cover, such as sector-year or
temporal-neighborhood covers.  The important point is that there is no single
unconditional global average over all firms, papers, reviews, or agents.
Globality is always relative to a declared cover, and non-gluing at a coarser
cover is a meaningful result.

\paragraph{Localized interventions.}
\textsc{Prometheus} treats interventions as local tests.  A \(j\)-do query
\(\doop_j(X=x)\) modifies histories or tests inside a context \(U\) and asks how
the local predictive-state table changes under comparable covers.  The result
is not automatically an identified causal effect in Pearl's sense
\citep{pearl2009causality}; it is an intervention-conditioned probe of the
language-derived world model.  \textsc{Prometheus} reports the support and provenance
behind the probe rather than presenting it as a source-free causal estimate.

More explicitly, let
\[
  j(U)=\{u_i:U_i\to U\}
\]
be a cover of contexts considered comparable for the query, and let
\[
  I^{a}_{U_i}:\Psh(U_i)\to \Psh^{\doop(a)}(U_i)
\]
be a local intervention map that edits a test, fixes an action, inserts a repair
step, or conditions on an explicitly declared regime.  The \(j\)-localized
intervention state is computed by restriction, local intervention, and
aggregation:
\[
  \doop_j(a)_U(s)
  =
  \operatorname{Agg}_{u_i:U_i\to U\in j(U)}
  \left(
    I^{a}_{U_i}\bigl(\rho_{U,U_i}(s)\bigr)
  \right).
\]
Compatibility is then checked after the intervention.  If the intervened local
sections glue, the atlas may report a coherent intervention-conditioned
prediction over \(U\).  If they do not, the query is only locally supported, and
the failed overlaps identify where comparability, measurement, or evidence
breaks down.

\section{The Claims Atlas}

The primary user-facing object is the Claims Atlas.  It is designed to answer
research questions that flat summaries obscure.

\paragraph{Main causal spine.}
The atlas extracts recurrent, high-support causal paths that organize the
corpus.  In an ocean-warming corpus, a spine may include warming, stratification,
oxygen loss, prey availability, migration, recruitment, and population change.
In SEC workflows, a spine may include investment, supply-chain constraints,
margin pressure, capital allocation, and realized outcomes.

\paragraph{Local context regions.}
Each spine is decomposed into local regions.  A region may correspond to a
species group, geography, time period, document cluster, product aspect, or
workflow stage.  Users can enter a region and inspect its local PSR, support,
claims, and provenance.

\paragraph{Drift detection.}
When local models change across time, retrieval runs, or document strata,
\textsc{Prometheus} reports drift.  Drift can be textual, causal, predictive, or
topological: the support distribution changes, a causal polarity changes, a
test prediction changes, or the overlap graph itself changes.

\paragraph{Regime tensions.}
The atlas highlights where local models resist gluing.  Some tensions are
contradictions; others are legitimate regime boundaries.  The interface should
make this distinction visible by exposing modifiers, populations, measurement
protocols, and source provenance.

\paragraph{Provenance drill-downs.}
Every atlas claim points back to evidence units.  A user can inspect source
passages, extracted rows, normalized claims, support counts, and neighboring
contexts.  Provenance is not decorative metadata; it is the mechanism by which
the atlas remains corrigible.

\section{Ocean-Temperature Artifact Case Study}
\label{sec:ocean-artifact}

We use a \textsc{Prometheus} GUI run on the query ``analyze 10 recent studies of the
impact of rising ocean temperatures on fish populations'' as the paper's
concrete artifact case study.  The run retrieved and acquired eleven documents
because the acquisition layer retained an additional closely related study.  The
corpus includes studies on marine fish endoparasites, larval thermal tolerance,
the North Atlantic heat wave, Argo temperature artifacts, kuruma shrimp
aquaculture, global fisheries economics, coral mortality, sea-grass carbon
storage, mixotrophic phytoplankton, and \emph{Vibrio vulnificus}.  This is a
good stress test for \textsc{Prometheus} because the query is nominally about fish
populations, but the retrieved literature naturally fans out into ecological,
measurement, aquaculture, microbial, economic, and carbon-cycle regimes.

The run produced \(3{,}065\) extracted events, \(11\) causal episodes, \(199\)
local contexts, \(199\) local PSRs, \(199\) sheaf objects, \(198\) restriction
and gluing diagnostics, \(160\) compatible restrictions, \(194\) compatible
gluing overlaps, and \(4\) tense gluing overlaps.  The LLM backend made
\(4{,}383\) requests using \(2{,}361{,}749\) total tokens, with an estimated API
cost of about \(\$1.24\).  The most frequent causal
relations in the extracted atlas were \emph{leads to} (\(1{,}152\)),
\emph{reduces} (\(577\)), \emph{increases} (\(466\)), \emph{influences}
(\(441\)), \emph{causes} (\(214\)), and \emph{affects} (\(214\)).  Prominent
high-support local contexts included larval survival under food scarcity,
subpolar gyre weakening, rising ocean surface temperatures, fisheries output
and secondary activities, mixotrophic metabolic evolution, zooxanthellae
density fluctuations, parasite transmission, and sea-grass habitat suitability.

\begin{table}[t]
\centering
\small
\begin{tabular}{lr}
\toprule
Artifact statistic & Value \\
\midrule
Acquired studies & 11 \\
Extracted events & 3,065 \\
Causal episodes & 11 \\
Local contexts / PSRs / sheaf objects & 199 / 199 / 199 \\
Restriction and gluing diagnostics & 198 \\
Compatible restrictions & 160 \\
Compatible gluing overlaps & 194 \\
Tense gluing overlaps & 4 \\
Topos world-model glue term & 3.5168 \\
LLM requests / tokens & 4,383 / 2,361,749 \\
Estimated LLM API cost & about \$1.24 \\
\bottomrule
\end{tabular}
\caption{Summary statistics for the ocean-temperature \textsc{Prometheus} artifact.}
\label{tab:ocean-artifact-stats}
\end{table}

\paragraph{\textsc{Prometheus} PSR bundle.}
The PSR bundle is the computational layer beneath the Claims Atlas.  It exposes
the finite predictive-state family rather than only the extracted causal rows:
the corpus-level PSR has rank \(2{,}784\), \(2{,}896\) histories, and
\(2{,}903\) tests, while the local family contains \(199\) local PSRs.  The
bundle reports \(160\) compatible restriction arrows out of \(198\) root
restriction checks, mean gluing loss \(0.0179\), no learned non-root overlap
edges, and no attached \(j\)-do probes for this particular run.  The absence of
\(j\)-do probes is itself informative: this artifact is a literature atlas, so
the main signal is not a computed intervention query but the support, drift,
and gluing behavior of local causal charts.

Table~\ref{tab:ocean-psr-contexts} shows representative local PSRs.  These are
the objects a researcher sees before reading source passages.  Each local PSR
has a finite set of histories and tests induced by extracted causal observation
events.  The restriction columns compare the local chart with the corpus chart;
the gluing columns show whether the local latent section is geometrically
compatible with the corpus section after projection.

\begin{table}[t]
\centering
\scriptsize
\begin{tabular}{p{0.23\linewidth}rrrrrrrp{0.16\linewidth}}
\toprule
Local context & Events & Rank & Hist. & Tests & Shared & Mean gap & Weighted loss & Interpretation \\
\midrule
Larval survival affected by food scarcity & 46 & 44 & 45 & 45 & 2,025 & 0.0219 & 0.001950 &
Core food-limitation and thermal-tolerance mechanism. \\
Subpolar gyre weakening effects & 44 & 37 & 41 & 40 & 1,640 & 0.0246 & 0.003436 &
Ocean-circulation route into North Atlantic warming. \\
Rising ocean surface temperatures & 38 & 24 & 30 & 30 & 900 & 0.0330 & 0.002639 &
Selected persistent-state focus context. \\
Fisheries output and secondary activities & 38 & 38 & 38 & 38 & 1,444 & 0.0260 & 0.004572 &
Economic value-chain consequences of fisheries output. \\
Metabolic evolution of mixotrophs & 37 & 30 & 33 & 33 & 1,089 & 0.0299 & 0.002163 &
Trophic feedback through heterotrophy and grazing pressure. \\
Sea-grass habitat suitability & 31 & 31 & 31 & 31 & 961 & 0.0319 & -- &
Neighboring carbon-storage and habitat-suitability regime. \\
\emph{Vibrio vulnificus} inflammatory response & 32 & 32 & 32 & 32 & 1,024 & 0.0309 & 0.002734 &
Marine disease and inflammatory-response neighbor. \\
Argo oceanographic float data corrections & 30 & 30 & 30 & 30 & 900 & 0.0330 & 0.002865 &
Measurement regime for correcting temperature records. \\
Climate change and maternal health & 29 & 25 & 27 & 27 & 729 & 0.0367 & -- &
Off-query health drift retained as a separate chart. \\
\bottomrule
\end{tabular}
\caption{Representative local PSRs from the ocean-temperature bundle.  ``Hist.''
denotes finite histories.  Shared cells and mean gap are corpus-to-context
restriction diagnostics.  Weighted loss is the gluing diagnostic when reported
in the bundle view.  The table shows why the artifact is not a single graph:
biological mechanisms, circulation mechanisms, economic consequences,
measurement issues, disease pathways, and retrieval drift all become separate
local predictive-state charts.}
\label{tab:ocean-psr-contexts}
\end{table}

The focus-context matrix makes the PSR concrete.  For the persistent focus
\emph{rising ocean surface temperatures}, the displayed local table has \(30\)
histories by \(30\) tests and rank \(24\).  Representative histories include
surface temperatures affecting evolutionary shifts back toward photosynthesis,
causing higher grazing rates that reduce prey abundance, increasing grazing
rates that reduce prey abundance, and increasing photosynthetic reliance over
heterotrophy.  Representative tests include the same prey-abundance mechanisms
plus a coastal-ecosystem-dynamics claim.  Most displayed cells have predictive
mass \(0.0312\), while two salient transitions have mass \(0.0938\): from the
history ``surface temperatures cause higher grazing rates which reduce prey
abundance'' to the corresponding grazing-rate test, and from the history
``surface temperatures increase grazing rates which reduce prey abundance'' to
the coastal-ecosystem-dynamics test.  Thus the local PSR does not merely store
claim counts; it records which histories make which causal tests more probable
inside a local chart.

The gluing-backpropagation rows provide examples of what the system treats as
stable local sections.  The larval-survival context has \(45\) overlap sections,
confidence \(0.9981\), weighted loss \(0.001950\), and maximum section gap
\(0.002810\); its representative sections include food scarcity reducing sea
urchin larval thermal tolerance, decreasing thermal tolerance, and increasing
physiological stress.  The subpolar-gyre context has \(41\) overlap sections,
confidence \(0.9966\), weighted loss \(0.003436\), and maximum section gap
\(0.004424\); its sections connect reduced Arctic-water inflow and warm
subtropical inflow to elevated North Atlantic temperatures.  The mixotroph
context has \(33\) overlap sections, confidence \(0.9978\), weighted loss
\(0.002163\), and maximum section gap \(0.005492\); it connects greater
heterotrophic reliance to higher grazing rates and reduced prey abundance.  The
focus context itself has \(30\) overlap sections, confidence \(0.9974\),
weighted loss \(0.002639\), and maximum section gap \(0.003835\), with sections
for reduced prey abundance and higher grazing rates.  These rows are the
mechanical reason the persistent state can recommend continuing from the
surface-temperature chart.

Drift appears in two complementary ways.  First, the Claims Atlas groups
off-query material into drift regions rather than suppressing it.  The
human-health drift region contains \(160\) events over \(11\) contexts,
including climate change and maternal health, malaria transmission and
pregnancy risks, healthcare access, water insecurity, and food insecurity.  The
off-query climate drift region contains \(58\) events over \(7\) contexts,
including Antarctic resource claims, treaty-meeting opposition, and Indian
Antarctic research-base proposals.  Second, the PSR bundle marks narrow
restriction or gluing rows as divergent when a local chart has too little
overlap or too large a gap to be transported.  Examples include
\emph{rising ocean temperatures} itself as a one-section gluing divergence
\((\)weighted loss \(0.1108)\), geopolitical tensions over Antarctic territory
\((0.1012)\), extreme weather and maternal health \((0.0915)\), and sea-grass
habitat-suitability shifts \((0.0824)\).  These rows are useful precisely
because they prevent the paper's example from pretending that broad retrieval
produced a clean, single-topic corpus.

The Claims Atlas is the paper-facing compression of this run.  Its first
screen does not present a single answer to the query.  Instead, it partitions
the extracted claims into named causal regions, identifies the recurrent causal
spine, and marks the places where the run drifted away from the requested
fish-population question.  The displayed atlas contains \(11\) documents,
\(3{,}065\) claims, \(200\) displayed local contexts including the corpus-level
view, \(152\) regime surfaces, \(4\) tense gluing overlaps, and mean glue loss
\(0.0179\).  The main atlas regions are summarized in
Table~\ref{tab:ocean-atlas-regions}.

\begin{table}[t]
\centering
\small
\begin{tabular}{p{0.28\linewidth}rrp{0.42\linewidth}}
\toprule
Atlas region & Events & Contexts & Interpretation \\
\midrule
Core query spine & 1,335 & 86 &
Fish-population, thermal-stress, aquaculture, parasite, warming, and survival
claims. \\
Other local contexts & 535 & 39 &
Residual climate and marine-health contexts not yet assigned to a named lens. \\
Neighboring marine regimes & 483 & 29 &
Coral, sea-grass, \emph{Vibrio}, and other marine systems adjacent to the
central fish-population target. \\
Marine ecosystem mechanisms & 303 & 14 &
Current, nutrient, prey, trophic, and metabolic mechanisms connecting warming
to ecosystem response. \\
Off-query health drift & 160 & 11 &
Human-health, pregnancy, malaria, water-security, and adaptation material
introduced by broad climate-temperature retrieval. \\
Economic consequences & 106 & 6 &
Fisheries value-chain and regional multiplier claims downstream of biological
population changes. \\
Measurement and data regimes & 85 & 6 &
Argo float, calibration, correction, and measurement-artifact claims that
affect the interpretation of warming signals. \\
Off-query climate drift & 58 & 7 &
Antarctic governance and resource-politics claims weakly connected to the
fish-population query. \\
\bottomrule
\end{tabular}
\caption{Claims Atlas partition for the ocean-temperature run.  The atlas makes
both central evidence and retrieval drift visible: most events sit in the core
query spine, but sizable neighboring, measurement, economic, and off-query
regions are preserved rather than averaged into one global summary.}
\label{tab:ocean-atlas-regions}
\end{table}

The highest-support spine claims show how the atlas differs from a generic
summary.  The most supported claim family is a larval-temperature mechanism:
food limitation reduces the thermal tolerance of purple sea urchin larvae by
lowering their capacity to maintain membrane fluidity and survive heat exposure
\((45\) extracted claims, \(8\) regime aliases).  A second family connects the
weakening of the subpolar gyre to increased warm subtropical inflow, elevated
North Atlantic temperatures, and changes in long-term ocean-temperature records
\((26\) claims, \(8\) aliases).  Other spine families connect rising surface
temperatures to mixotrophic phytoplankton trade-offs, increased grazing
pressure, lower prey abundance, endoparasite exposure through host feeding
behavior, and fisheries value-chain effects.  These are not merged into a
single ``warming harms fish'' proposition.  They are kept as local claim
families with documents, surface variants, regime aliases, and evidence
statements attached.

The atlas also exposes the run's regime tensions.  One obstruction involves the
claim that reduced larval survival lowers recruitment and affects kelp-forest
productivity: this link appears across kelp-forest and marine-heatwave regimes
with competing \emph{affects} and \emph{reduces} surfaces.  Several
regime-sensitive rows involve host feeding behavior and endoparasite diversity,
where the same subject-object pair appears across parasite-prevalence,
environmental-factor, and feeding-behavior contexts with different relation
surfaces.  Additional tensions appear around coral sedimentation stress,
\emph{Vibrio} virulence activation, and fisheries output multipliers.  The
point is not that these rows are errors.  They are exactly the places where a
researcher should inspect modifiers, measurement conditions, population, and
provenance before transporting a claim.

\begin{table}[t]
\centering
\scriptsize
\begin{tabular}{p{0.24\linewidth}p{0.24\linewidth}p{0.14\linewidth}p{0.24\linewidth}}
\toprule
Subject & Object & Atlas state & Regime interpretation \\
\midrule
Reduced survival rates in larvae & Kelp-forest productivity & Obstructed &
The larval-recruitment mechanism touches kelp-forest productivity through both
marine-heatwave and kelp-ecosystem regimes, so the atlas withholds a clean
global merge. \\
Elevated sedimentation & Physical and physiological stress on corals &
Regime-sensitive & Coral mortality, tissue-extraction, and environmental-stress
contexts carry related but non-identical surfaces. \\
Host feeding behavior & Specificity and diversity of endoparasite infections &
Regime-sensitive & Parasite-prevalence, environmental-factor, and
feeding-behavior contexts agree on exposure but vary in relation surface. \\
Activation of \emph{Vibrio} virulence factors & Disease progression in fish and
humans & Regime-sensitive & Bacterial virulence, iron availability, fish-human
transmission, and warming contexts meet but should remain locally annotated. \\
Direct output of marine capture fisheries & Secondary economic activities &
Regime-sensitive & Fisheries-output, value-chain, and multiplier contexts
transport only with economic-model assumptions. \\
Food limitation & Thermal tolerance of purple sea urchin larvae &
Multi-regime glued & Multiple larval-survival and food-limitation regimes
support a stable mechanism linking food scarcity to lower thermal tolerance. \\
\bottomrule
\end{tabular}
\caption{Examples of regime tensions surfaced by the Claims Atlas.  The
important feature is not only whether a row is globally consistent, but whether
the atlas can tell the researcher which local regimes make the claim portable,
which require caveats, and which should remain obstructed.}
\label{tab:ocean-regime-tensions}
\end{table}

The Persistent World State turns this atlas into a continuation state for
research.  The selected focus context is \emph{rising ocean surface
temperatures}.  Its local PSR has \(24\) effective local observations over a
\(30 \times 30\) history-test table.  Representative histories and tests include
surface temperatures affecting evolutionary shifts back toward photosynthesis,
causing higher grazing rates that reduce prey abundance, increasing
photosynthetic reliance over heterotrophy, and influencing coastal ecosystem
dynamics.  The state records one relevant restriction check from the corpus to
the focus context; it is aligned, with \(900\) shared cells and maximum gap
\(0.0934\).  The corresponding gluing overlap is also aligned, with weighted
loss \(0.0026\) over \(30\) sections.  The state therefore recommends accepting
the current world state for this focus: no blocking local issue was detected,
and no focus-context repair probes were attached.  In paper terms, the
persistent state is the system's explicit answer to ``where should the next
research step start?'' rather than a passive dashboard.

This example makes the central \textsc{Prometheus} claim concrete.  A conventional
summary would likely report that warming affects marine organisms through heat
stress, food limitation, disease, habitat shifts, and economic consequences.
The \textsc{Prometheus} artifact instead exposes the local structure behind that answer:
which studies support each region, which contexts glue cleanly to the corpus
model, which narrow regions remain tense, and which provenance paths justify a
claim.  The output is therefore not just a summary of recent literature.  It is
a persistent causal atlas that a researcher can inspect, revise, and extend.

\section{GLP-1 Weight-Loss Literature Case Study}
\label{sec:glp1-artifact}

We also ran \textsc{Prometheus} on the query ``Analyze 10 recent studies of the
weight loss drug GLP-1 and synthesize their joint support.''  The acquisition
layer retained \(11\) documents.  The selected corpus includes work on a MOGAT2
inhibitor that increases GLP-1 concentrations in obese mice, systematic and
network-review material on GLP-1 receptor agonists and co-agonists for adults
without diabetes, pharmacist counseling and misuse concerns, chronic kidney
disease, incretin polyagonists as bariatric-surgery alternatives,
drug-target Mendelian randomization for gastrointestinal outcomes,
pharmacovigilance studies of endocrine and dermatologic safety, tirzepatide and
cocaine motivation in rodents, semaglutide approval for MASH, and oral GLP-1
drug development.  This is a useful health-literature stress test because the
retrieval is not a clean efficacy-only corpus.  It mixes weight-loss efficacy,
safety signals, access and adherence, cardiometabolic benefit, drug-delivery
constraints, and off-query neighboring mechanisms.

\begin{table}[t]
\centering
\small
\begin{tabular}{lr}
\toprule
Artifact statistic & Value \\
\midrule
Acquired studies & 11 \\
Extracted events & 3,376 \\
Causal episodes & 11 \\
Local contexts / PSRs / sheaf objects & 191 / 191 / 191 \\
Restriction checks & 190 \\
Compatible restrictions & 149 \\
Divergent restrictions & 41 \\
Compatible gluing overlaps & 186 \\
Tense gluing overlaps & 4 \\
Mean gluing loss & 0.0189 \\
Topos world-model glue term & 3.4468 \\
Corpus PSR rank / histories / tests & 2,997 / 3,144 / 3,152 \\
\bottomrule
\end{tabular}
\caption{Summary statistics for the GLP-1 weight-loss \textsc{Prometheus}
artifact.  The corpus is intentionally heterogeneous: it contains direct
weight-loss efficacy material, safety and pharmacovigilance material,
implementation and access material, and adjacent metabolic-disease and
addiction studies.}
\label{tab:glp1-artifact-stats}
\end{table}

The most frequent relation families were \emph{leads to} (\(1{,}203\)),
\emph{reduces} (\(831\)), \emph{increases} (\(546\)), \emph{influences}
(\(489\)), \emph{affects} (\(176\)), and \emph{causes} (\(129\)).  These counts
already show why a flat answer is fragile: the corpus contains benefit claims,
mechanism claims, access claims, and adverse-event claims, and these should not
be collapsed into a single endorsement or warning.

\paragraph{\textsc{Prometheus} PSR bundle.}
The GLP-1 PSR bundle exposes the corpus as a health-research atlas rather than
as a single therapy summary.  The corpus-level PSR has rank \(2{,}997\),
\(3{,}144\) histories, and \(3{,}152\) tests.  There are no learned non-root
overlap edges and no attached \(j\)-do probes in this run; the signal is
therefore carried by root-to-context restrictions and gluing diagnostics.  The
largest local PSRs are safety, access, dual-incretin, counseling, and metabolic
mechanism charts, not only direct weight-loss-efficacy charts.  This is the
main practical difference between an atlas and a review abstract: the artifact
keeps the benefit and risk surfaces inspectable as separate local sections.

\begin{table}[t]
\centering
\scriptsize
\begin{tabular}{p{0.26\linewidth}rrrrrrp{0.20\linewidth}}
\toprule
Local context & Events & Rank & Hist. & Tests & Shared & Mean gap & Interpretation \\
\midrule
Telogen effluvium and androgenetic alopecia & 107 &
89 & 98 & 98 & 9,604 & 0.0099 &
Largest safety basin; rapid weight loss and agent-specific hair-loss signals. \\
Improved drug accessibility and storage & 75 &
73 & 74 & 74 & 5,476 & 0.0132 &
Oral/non-peptide delivery, adherence, and storage constraints. \\
Tirzepatide dual GLP-1/GIP agonism & 65 &
58 & 61 & 60 & 3,660 & 0.0163 &
Dual-incretin mechanism spanning appetite, metabolism, and reward pathways. \\
Alopecia and hair-loss events & 63 &
61 & 62 & 62 & 3,844 & 0.0158 &
Dermatologic adverse-event region with safety-specific provenance. \\
Altered GLP-1 secretion and response & 45 &
42 & 43 & 43 & 1,849 & 0.0229 &
Incretin signaling, adiposity, appetite, and energy-balance mechanism. \\
Obesity treatment pharmacotherapy & 43 &
30 & 35 & 34 & 1,190 & 0.0291 &
Central treatment region spanning efficacy, safety, and agent choice. \\
Counseling on GLP-1 receptor agonists & 43 &
40 & 41 & 41 & 1,681 & 0.0241 &
Practice-facing counseling, misuse, and patient-education chart. \\
Non-surgical alternatives to bariatric surgery & 42 &
36 & 39 & 39 & 1,521 & 0.0253 &
Polyagonists as a treatment-regime alternative to surgery. \\
Residual cardiorenal risk factors & 39 &
37 & 38 & 38 & 1,444 & 0.0260 &
CKD and cardiovascular-benefit neighboring outcome chart. \\
Improved glucose tolerance and insulin sensitivity & 37 &
37 & 37 & 37 & 1,369 & 0.0267 &
Mouse metabolic mechanism linked to MOGAT2 inhibition and GLP-1 levels. \\
\bottomrule
\end{tabular}
\caption{Representative local PSRs from the GLP-1 bundle.  ``Hist.'' denotes
finite histories.  Shared cells and mean gap are corpus-to-context restriction
diagnostics.  The high-support charts show that the run is not simply an
efficacy summary: safety, access, adherence, dual-agonist mechanisms,
cardiorenal outcomes, and animal metabolic mechanisms all become separate local
predictive-state charts.}
\label{tab:glp1-psr-contexts}
\end{table}

\begin{table}[t]
\centering
\small
\begin{tabular}{p{0.31\linewidth}rrp{0.42\linewidth}}
\toprule
Atlas lens & Events & Contexts & Interpretation \\
\midrule
Weight-loss and metabolic efficacy & 310 & 10 &
GLP-1 and co-agonist efficacy, weight-loss mechanisms, glucose tolerance,
insulin sensitivity, and lipid metabolism. \\
Safety and pharmacovigilance & 346 & 10 &
Hair loss, telogen effluvium, endocrine and reproductive signals,
dose dependence, and comparative agent safety. \\
Access, counseling, and adherence & 299 & 9 &
Oral delivery, storage, pharmacist knowledge, patient education, demand,
misuse, and safe-medication-use training. \\
Comorbidity and health-outcome neighbors & 293 & 10 &
CKD, residual cardiorenal risk, MASH, metabolic inflammation, fibrosis,
and cardiovascular event reduction. \\
Mechanism and alternative target discovery & 259 & 9 &
Dual agonism, incretin response, MOGAT2 inhibition, receptor signaling,
and alternative sustained-weight-loss mechanisms. \\
Off-query reward-pathway neighbor & 213 & 9 &
Tirzepatide, dopamine, cocaine self-administration, craving, and addiction
mechanisms retained as a separate neighboring chart. \\
GI causal-inference and mediation surface & 119 & 9 &
Mendelian randomization, paralytic ileus, BMI/HbA1c mediation, and
conflicting observational evidence. \\
\bottomrule
\end{tabular}
\caption{Paper-facing atlas lenses for the GLP-1 run.  Unlike the
ocean-temperature artifact, the automatic Claims Atlas placed almost all local
contexts in a residual bucket plus a tiny measurement bucket.  For exposition,
we therefore group local contexts into non-exclusive health-research lenses.
The non-exclusivity is itself informative: GLP-1 evidence crosses efficacy,
safety, access, comorbidity, mechanism, and off-query reward-pathway surfaces.}
\label{tab:glp1-atlas-regions}
\end{table}

The gluing diagnostics expose which parts of the corpus transport cleanly and
which require local caveats.  Large regions such as telogen effluvium and
androgenetic alopecia, improved drug accessibility and storage, tirzepatide
dual GLP-1/GIP agonism, obesity-treatment pharmacotherapy, patient education,
glucose-tolerance and insulin testing, and safety profiles of GLP-1 receptor
agonists align with the corpus section.  The persistent state nevertheless
marks the overall state as provisional because it finds \(41\) divergent
restriction checks and \(4\) divergent gluing overlaps.

\begin{table}[t]
\centering
\scriptsize
\begin{tabular}{p{0.24\linewidth}p{0.24\linewidth}p{0.14\linewidth}p{0.24\linewidth}}
\toprule
Subject & Object & Atlas state & Regime interpretation \\
\midrule
GLP-1RA signaling & Paralytic ileus risk & Obstructed &
Conflicting observational evidence and Mendelian-randomization regimes expose
different relation surfaces, so the atlas withholds a clean global merge. \\
Activation of GLP-1/GIP receptors & Cocaine seeking and relapse behaviors &
Obstructed & Reward-pathway material is mechanistically adjacent to tirzepatide
but off-query for obesity treatment and should remain locally annotated. \\
GLP-1RA signaling & Risk of paralytic ileus & Obstructed &
The same GI target appears with both \emph{affects} and \emph{reduces}
surfaces in a narrow evidence regime. \\
Increased GLP-1 receptor agonist signaling & Risk of paralytic ileus &
Multi-regime glued & Causal-inference, protective-effect, and conflicting
observational regimes agree on a protective direction after localization. \\
GLP-1 receptor agonist & Weight loss in overweight or obese adults without
diabetes & Multi-regime glued & Co-agonist, comparative agent, efficacy,
safety, and clinical-trial regimes support the central weight-loss effect. \\
GLP-1 receptor agonist & Metabolic stress and inflammation in CKD &
Multi-regime glued & CKD, cardiovascular, kidney-benefit, and inflammatory
pathway contexts support a neighboring cardiorenal-benefit chart. \\
GLP-1 receptor agonist & Significant weight loss in overweight or obese adults
without diabetes & Multi-regime glued & Efficacy and safety-assessment
regimes support the benefit claim while preserving population scope. \\
Reduced dopamine release & Decreased motivation to self-administer cocaine &
Multi-regime glued & Addiction and reward-pathway contexts glue internally,
but remain a neighboring regime rather than part of the weight-loss spine. \\
\bottomrule
\end{tabular}
\caption{Examples of regime tensions and glued surfaces in the GLP-1 Claims
Atlas.  The important behavior is selective transport: core efficacy and some
cardiorenal claims glue across regimes, while GI mediation, hair-loss,
hormonal, and reward-pathway claims remain narrow or obstructed.}
\label{tab:glp1-regime-tensions}
\end{table}

This is exactly the behavior wanted in a health setting.  \textsc{Prometheus}
does not convert heterogeneous literature into individualized medical advice or
a global causal verdict.  It preserves locality: weight-loss efficacy in
non-diabetic adults, mouse metabolic mechanisms, polyagonist alternatives,
cardiorenal benefits, MASH material, pharmacist counseling, pharmacovigilance
signals, and hair-loss/hormonal safety concerns are visible as different local
sections.  The artifact's conclusion is not ``GLP-1 works'' or ``GLP-1 is
risky.''  It is a navigable map of where the corpus supports treatment,
which mechanisms and populations carry that support, and where safety or
transportability claims should remain under inspection.

\section{Resveratrol and Red-Wine Health-Benefit Case Study}
\label{sec:resveratrol-artifact}

The third artifact case study uses the query ``Analyze 10 recent studies of the
health benefits of Resveratrol in red wine and synthesize their joint
support.''  The acquisition layer retained \(13\) documents.  The selected
corpus includes a 2025 systematic review of red-wine consumption and
cardiovascular risk, work on alcohol hypersensitivity in aspirin-exacerbated
respiratory disease, winemaking-residue valorization, older studies on
red-wine resveratrol stability and high-trans-resveratrol wine, a trauma-
hemorrhage organ-function study, a review of red wine and resveratrol effects
on human health, intestinal-cancer material, commentary on resveratrol,
microbiota-derived resveratrol metabolites as biomarkers of red-wine
consumption, and a Mediterranean-diet well-being study.  This corpus is a good
test of \textsc{Prometheus} because the phrase ``health benefits of resveratrol
in red wine'' pulls together benefit claims, wine-processing claims,
bioavailability objections, biomarker studies, cell-line and animal studies,
and diet-measurement drift.

\begin{table}[t]
\centering
\small
\begin{tabular}{lr}
\toprule
Artifact statistic & Value \\
\midrule
Acquired studies & 13 \\
Extracted events & 4,057 \\
Causal episodes & 13 \\
Local contexts / PSRs / sheaf objects & 227 / 227 / 227 \\
Restriction checks & 226 \\
Compatible restrictions & 177 \\
Divergent restrictions & 49 \\
Compatible gluing overlaps & 221 \\
Tense gluing overlaps & 5 \\
Mean gluing loss & 0.0178 \\
Topos world-model glue term & 3.8582 \\
Corpus PSR rank / histories / tests & 3,632 / 3,802 / 3,813 \\
\bottomrule
\end{tabular}
\caption{Summary statistics for the Resveratrol/red-wine
\textsc{Prometheus} artifact.  The corpus contains direct cardiovascular and
anti-inflammatory claims, mechanistic SIRT1 and cancer-cell claims,
bioavailability concerns, wine-production and stability claims, biomarker
claims, and diet-measurement drift.}
\label{tab:resveratrol-artifact-stats}
\end{table}

The most frequent relation families were \emph{leads to} (\(1{,}439\)),
\emph{reduces} (\(827\)), \emph{increases} (\(727\)), \emph{influences}
(\(621\)), \emph{causes} (\(256\)), and \emph{affects} (\(182\)).  The top
spine claims include resveratrol reducing oxidative stress in cardiovascular
tissues, poor bioavailability limiting systemic effects, SIRT1 activation
influencing glucose metabolism and PCOS-related mechanisms, red-wine
polyphenols improving lipid profile and endothelial function, cyclooxygenase
inhibition in Caco-2 cell contexts, and microbiota-derived metabolite links to
inflammation.

\paragraph{\textsc{Prometheus} PSR bundle.}
The Resveratrol PSR bundle has a corpus-level rank of \(3{,}632\), with
\(3{,}802\) histories and \(3{,}813\) tests.  The largest local charts are not
all clinical-benefit charts.  They include resveratrol metabolism and cancer
prevention, Caco-2 cell models, grape-extract intervention, lipid metabolism,
fermentation temperature, food-frequency-questionnaire limitations, and
anti-inflammatory effects.  The bundle therefore surfaces a central limitation
of the red-wine/resveratrol literature: transportability depends on whether the
claim is about wine chemistry, consumed red wine, resveratrol as an isolated
compound, microbial metabolites, a cell-line assay, or human cardiovascular
risk.

\begin{table}[t]
\centering
\scriptsize
\begin{tabular}{p{0.26\linewidth}rrrrrrp{0.20\linewidth}}
\toprule
Local context & Events & Rank & Hist. & Tests & Shared & Mean gap & Interpretation \\
\midrule
Resveratrol metabolism and cancer prevention & 78 &
57 & 67 & 66 & 4,422 & 0.0149 &
Largest mechanistic chart, joining metabolism, cancer prevention, and
bioactive-compound pathways. \\
Caco-2 human colonic adenocarcinoma cells & 75 &
53 & 62 & 62 & 3,844 & 0.0159 &
Cell-line chart for intestinal-cancer and differentiation claims. \\
Grape extract dietary intervention & 70 &
44 & 54 & 53 & 2,862 & 0.0186 &
Dietary grape-extract and tumor-reduction mechanisms. \\
Resveratrol's role in lipid metabolism & 59 &
53 & 56 & 56 & 3,136 & 0.0176 &
Lipid, endothelial, and oxidative-stress mechanisms. \\
Fermentation temperature effects & 56 &
51 & 53 & 53 & 2,809 & 0.0186 &
Wine-production chart affecting trans-resveratrol content and stability. \\
Caco-2 cell proliferation inhibition & 50 &
40 & 45 & 45 & 2,025 & 0.0219 &
Cyclooxygenase, viability, and cancer-cell-growth mechanism. \\
Limitations of food frequency questionnaires & 48 &
46 & 47 & 47 & 2,209 & 0.0210 &
Measurement chart for dietary intake and Mediterranean-diet evidence. \\
Resveratrol's anti-inflammatory effects & 44 &
38 & 41 & 41 & 1,681 & 0.0241 &
Inflammation, PI3K/Akt/MAPK, and organ-function mechanisms. \\
Polyphenols in red wine and heart health & 42 &
35 & 38 & 38 & 1,444 & 0.0260 &
Cardiovascular-benefit and endothelial-function chart. \\
Scientific controversy and data falsification & 17 &
17 & 17 & 17 & 289 & 0.0585 &
Credibility and controversy chart that limits naive benefit synthesis. \\
\bottomrule
\end{tabular}
\caption{Representative local PSRs from the Resveratrol bundle.  ``Hist.''
denotes finite histories.  Shared cells and mean gap are corpus-to-context
restriction diagnostics.  The table shows why the artifact is not a single
``red wine is healthy'' graph: production chemistry, cell models,
bioavailability, cardiovascular mechanisms, inflammation, measurement, and
controversy become separate predictive-state charts.}
\label{tab:resveratrol-psr-contexts}
\end{table}

\begin{table}[t]
\centering
\small
\begin{tabular}{p{0.31\linewidth}rrp{0.42\linewidth}}
\toprule
Atlas lens & Events & Contexts & Interpretation \\
\midrule
Cardiovascular and lipid-benefit spine & 168 & 5 &
French-paradox, red-wine cardiovascular risk, polyphenols, lipid profile,
endothelial function, and heart-health mechanisms. \\
Oxidative-stress and inflammation mechanisms & 188 & 5 &
Antioxidant, anti-inflammatory, PI3K/Akt/MAPK, organ-function, and
inflammation-reduction pathways. \\
Bioavailability, SIRT1, fertility, and ageing & 196 & 6 &
Poor bioavailability, rapid hepatic degradation, SIRT1 activation,
glucose metabolism, PCOS, infertility, and ageing claims. \\
Cancer and cell-line evidence & 237 & 5 &
Caco-2 cell proliferation, colonic adenocarcinoma, cyclooxygenase inhibition,
grape-extract dietary intervention, and tumor-reduction mechanisms. \\
Wine chemistry and production stability & 176 & 6 &
Fermentation temperature, storage, grape variety, stems, extraction,
trans-resveratrol, and cis--trans stability. \\
Microbiota, biomarkers, and diet measurement & 141 & 4 &
Microbiota-derived metabolites, urinary biomarkers, FFQ limitations,
Mediterranean-diet adherence, and emotional-well-being drift. \\
Controversy and hypersensitivity surface & 90 & 4 &
Scientific controversy, data-falsification concerns, basophil/eosinophil
activation, alcohol sensitivity, and AERD mechanisms. \\
\bottomrule
\end{tabular}
\caption{Paper-facing atlas lenses for the Resveratrol run.  The generated
Claims Atlas again placed most contexts in a residual bucket, so the paper
groups local contexts into non-exclusive research lenses.  The point is not to
force one global health claim, but to show the distinct evidence surfaces that
must be inspected before transporting any benefit claim.}
\label{tab:resveratrol-atlas-regions}
\end{table}

The persistent state marks the Resveratrol artifact as provisional: the corpus
has \(49\) divergent restriction checks and \(5\) divergent gluing overlaps.
The tense gluing rows are concentrated in narrow but epistemically important
regions: cardiovascular benefits of moderate red wine, red/white wine
concentration differences, inflammation reduction via resveratrol metabolites,
alcohol-sensitivity mechanisms in chronic rhinosinusitis with nasal polyps, and
urinary biomarkers of red-wine intake.  These are exactly the places where the
literature should not be flattened into one answer.

\begin{table}[t]
\centering
\scriptsize
\begin{tabular}{p{0.24\linewidth}p{0.24\linewidth}p{0.14\linewidth}p{0.24\linewidth}}
\toprule
Subject & Object & Atlas state & Regime interpretation \\
\midrule
Light exposure & Stability of trans-resveratrol & Obstructed &
Stability and isomerization regimes carry both \emph{affects} and
\emph{reduces} surfaces, so wine-storage claims do not glue cleanly. \\
Oak barrel aging & Concentration of trans and cis resveratrol in wine &
Obstructed & Production and aging conditions change the compound-concentration
surface and should not be merged with health-effect claims. \\
Resveratrol intake & Activation of the SIRT1 enzyme & Regime-sensitive &
Mechanism, PCOS, fertility, ageing, and metabolic-benefit regimes use both
\emph{causes} and \emph{increases} surfaces. \\
Reduction in oxidative stress & Decreased inflammation and cellular damage &
Regime-sensitive & Antioxidant and cardiovascular regimes agree directionally
but differ in whether oxidative-stress reduction is a mediator or endpoint. \\
Activation of the PI3K/Akt pathway & Improved cardiac function &
Regime-sensitive & Trauma-hemorrhage, anti-inflammatory, and signaling-pathway
contexts make this a local mechanism rather than a general red-wine claim. \\
Muscat Bailey A grape variety & Higher trans-resveratrol production &
Regime-sensitive & Grape-variety, sun-exposure, and high-resveratrol wine
contexts affect production claims, not clinical benefit directly. \\
Activation of eosinophils & Degranulation & Regime-sensitive &
AERD hypersensitivity material is an adverse-response neighbor to the red-wine
benefit question. \\
Addition of grape fruit stems during fermentation & Trans-resveratrol content &
Regime-sensitive & Fermentation and grape-stem regimes affect resveratrol
content and therefore modify, but do not prove, downstream health claims. \\
\bottomrule
\end{tabular}
\caption{Examples of regime tensions surfaced by the Resveratrol Claims Atlas.
The artifact distinguishes benefit mechanisms from wine-production chemistry,
bioavailability constraints, cell-line evidence, hypersensitivity mechanisms,
and measurement or controversy surfaces.}
\label{tab:resveratrol-regime-tensions}
\end{table}

This case study shows why the atlas representation matters for nutrition and
natural-product claims.  A flat summary would be tempted to say that
resveratrol in red wine is antioxidant and cardioprotective.  The
\textsc{Prometheus} artifact instead says something more useful: some local
contexts support antioxidant, lipid, endothelial, inflammatory, and cell-line
mechanisms; other contexts warn that bioavailability, wine chemistry,
measurement, hypersensitivity, and credibility issues determine whether the
claim transports.  The resulting object is not advice to drink red wine.  It is
a map of which parts of the literature support which mechanistic claims, and
where those claims should remain local.

\section{Additional Case-Study Templates}

We describe five case-study templates that exercise different parts of the
\textsc{Prometheus} contract.

\paragraph{Ocean warming and fish populations.}
The corpus consists of papers and reports on ocean warming, oxygen loss,
stratification, prey shifts, habitat migration, reproductive success, and fish
population dynamics.  The atlas should reveal the main causal spine and show
which species, geographies, and time scales support or break each link.

\paragraph{Product reviews.}
Shoes-ACOSI and targeted-sentiment datasets provide product-feedback corpora in
which local contexts include fit, comfort, activity, failure mode, return risk,
service, price/value, and quality.  \textsc{Prometheus} builds local review-experience
PSRs and exposes gluing tensions such as ``comfortable'' in short-use contexts
versus ``painful'' in long-mileage contexts.

\paragraph{SEC workflows.}
Annual filings and earnings-call transcripts describe plans, risks, investments,
and outcomes.  Contexts include company-year, sector, strategic theme, and
workflow stage.  The atlas can compare claims about digital investment,
supply-chain optimization, price actions, margin effects, and risk exposure
across years.

\paragraph{Health literature.}
Health corpora stress the need for locality.  Population, dosage, study design,
endpoint, and time horizon often determine whether claims transport.  A
\textsc{Prometheus} atlas should make non-transportability visible and should never
present a local literature claim as individualized medical advice.

\paragraph{Network-economy and agent traces.}
In small simulations, producers, transporters, and consumers emit local textual
reports.  \textsc{Prometheus} converts these reports into role-local PSRs, then measures
whether supply, capacity, demand, and payoff contexts glue into a coherent
operating regime.  These traces test whether the same atlas machinery can serve
agentic decision support.

\section{Grounded Counterfactuals with Paper Source Code}
\label{sec:grounded-counterfactuals}

The previous case studies treat counterfactuals as probes of the
language-derived world model.  This is useful, but it is still model-internal:
the atlas can ask how a local causal neighborhood would change if a claim,
mechanism, or repair were edited, but the result is only as grounded as the
extracted claims and their support.  A stronger opportunity appears when a
scientific paper ships source data, executable code, simulation inputs, or
plot-generation artifacts.  In that setting, \textsc{Prometheus} can bind a
symbolic sheaf intervention to an external scientific substrate, execute the
intervention there, and then push the measured result back into the topos world
model.

We tested this mode in four separate domains.  The first is a climate-forcing
example from airborne microplastics, where source tables and gridded figure
data support a direct optical-forcing intervention.  The second is a
palaeohydrology example from the Indus Valley Civilization, where
VIC-derived discharge anomalies, climate figure data, and an accompanying VIC
codebase support a drought-restoration intervention.  The third is the
well-known Sachs protein-signaling benchmark, where single-cell perturbation
data support measured experimental-regime substitutions.  The fourth is a
comparative-neuroscience example from singing mice, where MAPseq projection
matrices support a species-level intervention on motor-cortex projection
expansion.  The point is not that the same numeric machinery applies
everywhere.  The point is that the same Topos World Model loop recurs: extract
a claim sheaf, locate an external scientific substrate, execute or evaluate an
intervention there, and rebuild the local world model from the changed
observations.

\subsection{Microplastics: Optical-Forcing Intervention}

We first tested this mode on the Nature Climate Change paper \emph{Atmospheric
warming contributions from airborne microplastics and nanoplastics}
\citep{liu2026atmosphericMicroplastics}.  The paper reports that atmospheric
microplastic and nanoplastic particles (MNPs) have a mean direct radiative
forcing of \(0.039 \pm 0.019\,\mathrm{W\,m^{-2}}\), equivalent to about
\(16.2\%\) of black-carbon forcing, and that colored particles absorb much more
strongly than pristine particles.  The local paper folder contains both the PDF
and the source-data spreadsheets used for the reported figures, including the
workbook behind the paper's gridded forcing map and source tables for
microplastic and nanoplastic optical-forcing values.  This makes it possible
to evaluate a counterfactual that is not merely verbal:
\[
  \text{colored MNP optical forcing}
  \longmapsto
  \text{white/pristine MNP optical forcing}.
\]

\paragraph{Language topos.}
We first built a \textsc{Prometheus} claims world model from the document-level
causal extraction.  The baseline paper topos contained \(11\) extracted causal
events, \(8\) local PSRs, \(8\) sheaf objects, \(7\) root restriction checks,
and \(7\) gluing diagnostics.  The extracted local contexts included colored
MNP light absorption, atmospheric ageing optical effects, direct MNP radiative
forcing, regional forcing hotspots, radiative-transfer estimation, and the
interpretation of MNPs as previously unrecognized climate-forcing agents.  In
the ordinary sheaf-query layer, a counterfactual question about suppressing
colored-particle absorption would remain an internal claim-topos probe.

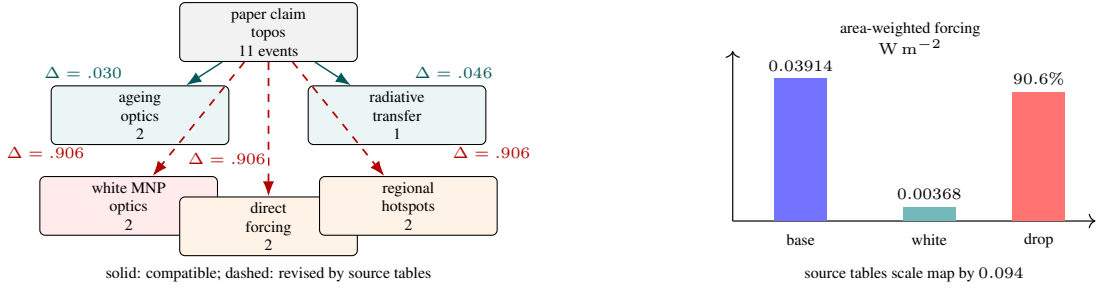
\begin{figure}[t]
\vspace{-0.6em}
\centering
\begin{minipage}[t]{0.48\linewidth}
\centering
\begin{tikzpicture}[
  ctx/.style={draw, rounded corners=2pt, align=center, font=\tiny, inner sep=2pt, minimum width=2.35cm, minimum height=0.50cm},
  good/.style={-{Latex[length=2mm]}, line width=0.55pt, draw=teal!70!black},
  tense/.style={-{Latex[length=2mm]}, line width=0.65pt, draw=red!70!black, dashed}
]
\node[ctx, fill=gray!10] (corpus) at (0,1.85) {paper claim\\topos\\11 events};
\node[ctx, fill=teal!8] (age) at (-1.7,0.75) {ageing\\optics\\2};
\node[ctx, fill=teal!8] (rt) at (1.7,0.75) {radiative\\transfer\\1};
\node[ctx, fill=red!8] (white) at (-1.85,-0.45) {white MNP\\optics\\2};
\node[ctx, fill=orange!10] (forcing) at (0,-0.72) {direct\\forcing\\2};
\node[ctx, fill=orange!10] (hot) at (1.85,-0.45) {regional\\hotspots\\2};
\draw[good] (corpus) -- (age);
\draw[good] (corpus) -- (rt);
\draw[tense] (corpus) -- (white);
\draw[tense] (corpus) -- (forcing);
\draw[tense] (corpus) -- (hot);
\node[font=\tiny, fill=white, inner sep=0.7pt, text=teal!70!black] at (-2.45,1.32) {$\Delta=.030$};
\node[font=\tiny, fill=white, inner sep=0.7pt, text=teal!70!black] at (2.45,1.32) {$\Delta=.046$};
\node[font=\tiny, fill=white, inner sep=0.7pt, text=red!70!black] at (-2.95,0.25) {$\Delta=.906$};
\node[font=\tiny, fill=white, inner sep=0.7pt, text=red!70!black] at (-0.55,0.18) {$\Delta=.906$};
\node[font=\tiny, fill=white, inner sep=0.7pt, text=red!70!black] at (2.95,0.25) {$\Delta=.906$};
\node[font=\tiny, align=left] at (0,-1.35) {solid: compatible; dashed: revised by source tables};
\end{tikzpicture}
\end{minipage}
\hfill
\begin{minipage}[t]{0.48\linewidth}
\centering
\begin{tikzpicture}[x=1cm,y=1cm]
\draw[->, line width=0.45pt] (0,0) -- (4.8,0);
\draw[->, line width=0.45pt] (0,0) -- (0,2.25);
\foreach \x/\lab in {0.9/base,2.6/white,4.05/drop} {
  \node[font=\tiny] at (\x,-0.25) {\lab};
}
\fill[blue!55] (0.55,0) rectangle (1.25,1.88);
\fill[teal!55] (2.25,0) rectangle (2.95,0.18);
\fill[red!55] (3.70,0) rectangle (4.40,1.70);
\node[font=\tiny] at (0.90,2.05) {$0.03914$};
\node[font=\tiny] at (2.60,0.36) {$0.00368$};
\node[font=\tiny] at (4.05,1.88) {$90.6\%$};
\node[font=\tiny, align=center] at (2.35,2.42) {area-weighted forcing\\\(\mathrm{W\,m^{-2}}\)};
\node[font=\tiny, align=center] at (2.40,-0.70) {source tables scale map by \(0.094\)};
\end{tikzpicture}
\end{minipage}
\vspace{-0.4em}
\caption{A concrete microplastics artifact slice.  Left: local contexts in the
counterfactual sheaf after replacing colored-particle optics with the
white/pristine source-table value.  Right: the executable source-data proxy
holds the published spatial distribution fixed and reduces the area-weighted
MNP forcing from \(0.03914\) to \(0.00368\,\mathrm{W\,m^{-2}}\).}
\label{fig:microplastics-artifact-sheaf}
\vspace{-0.8em}
\end{figure}

\paragraph{Executable intervention.}
We then used the paper's source spreadsheets as an executable substrate.  The
intervention holds the published spatial MNP distribution fixed and replaces
the colored-particle optical-forcing value with the white/pristine value
derived from the paper's Source Table~1 and Source Table~2.  Let
\(F_{\mathrm{base}}\) be the published gridded all-sky MNP direct-radiative
forcing map and let \(m_{\mathrm{base}}\) and \(m_{\mathrm{white}}\) be the
table-derived MP+NP mean forcing values.  The first executable proxy computes
\[
  F_{\mathrm{cf}}
  =
  F_{\mathrm{base}}
  \frac{m_{\mathrm{white}}}{m_{\mathrm{base}}}.
\]
In the artifact, \(m_{\mathrm{base}}=0.039\) and
\(m_{\mathrm{white}}=0.0036667\,\mathrm{W\,m^{-2}}\), so the scale factor is
\(0.0940\).  Area-weighting the gridded map by latitude gives
\[
  \bar F_{\mathrm{base}}=0.03914,\qquad
  \bar F_{\mathrm{cf}}=0.00368,\qquad
  \bar F_{\mathrm{base}}-\bar F_{\mathrm{cf}}=0.03546
  \quad \mathrm{W\,m^{-2}}.
\]
Thus the colored-to-white intervention suppresses about \(90.6\%\) of the
modeled MNP forcing in this source-table proxy.  Relative to the extracted
black-carbon benchmark map, the baseline MNP forcing is \(14.29\%\) of
black-carbon forcing by area-weighted mean, while the white-equivalent
counterfactual is about \(1.34\%\).  \Cref{fig:microplastics-artifact-sheaf}
shows the corresponding counterfactual sheaf slice and source-data proxy.

\paragraph{Back into the sheaf.}
The important step is not only computing the numeric proxy.  After execution,
\textsc{Prometheus} rewrites the relevant causal observations and rebuilds the
world model from the modified episode.  Six source claims are changed.  The
baseline observations
\begin{quote}
\footnotesize\ttfamily
colored\_mnp\_light\_absorption|increase|light\_absorption
\end{quote}
and related direct-forcing and hotspot claims are replaced by counterfactual
observations such as
\begin{quote}
\footnotesize\ttfamily
counterfactual\_white\_mnp\_optics|sets\_to|\\
white\_pristine\_absorption
\end{quote}
\begin{quote}
\footnotesize\ttfamily
counterfactual\_mnp\_direct\_radiative\_forcing|produces|\\
white\_equivalent\_mean\_direct\_radiative\_forcing
\end{quote}
and
\begin{quote}
\footnotesize\ttfamily
counterfactual\_regional\_mnp\_forcing\_hotspots|reduced\_by|\\
0.906\_forcing\_fraction.
\end{quote}
The modified episode is then passed through the ordinary Topos World Model
builder.  The resulting counterfactual world model again has \(11\) events,
\(8\) local PSRs, \(8\) sheaf objects, \(7\) restriction checks, and \(7\)
gluing diagnostics, but the local cover has changed: it now contains
\texttt{counterfactual\_white\_mnp\_optics},
\texttt{counterfactual\_mnp\_direct\_radiative\_forcing}, and
\texttt{counterfactual\_regional\_mnp\_forcing\_hotspots}.  The intervention is
therefore not a dashboard annotation.  It is a new sheaf world model.

\paragraph{Interface result.}
The regenerated sheaf explorer exposes this mode as a \emph{Grounded
Counterfactual Layer}.  Its top panel shows the intervention flow
\[
  \text{original sheaf section}
  \to
  \text{paper source tables and forcing-grid workbook}
  \to
  \text{modified sheaf section},
\]
alongside the measured forcing drop and the number of modified causal events.
Buttons jump directly into the modified local contexts, where the user can
inspect the local Hankel table, sheaf sections, and gluing diagnostics.  This
is the interface analogue of the formal point: grounded counterfactuals are
not only answers; they are new local worlds whose compatibility with the rest
of the atlas can be inspected.

\paragraph{Why this changes the \textsc{Prometheus} contract.}
This example separates two kinds of counterfactuals.  An \emph{internal}
counterfactual is evaluated only inside the extracted language topos.  It is a
useful research probe, but it remains symbolic.  A \emph{grounded}
counterfactual is evaluated against an external executable substrate supplied
by the paper, then reincorporated into the topos as changed causal
observations.  The resulting loop is
\[
  \text{text-derived claim topos}
  \to
  \text{executable source-data intervention}
  \to
  \text{measured effect}
  \to
  \text{rebuilt counterfactual sheaf}.
\]
For papers that include code, equations, tables, simulation outputs, or
plot-generation data, this gives \textsc{Prometheus} a new source of power:
counterfactual reasoning can be calibrated against a known scientific model
when one is available, while still preserving the local, corrigible, and
inspectable structure of the language-derived world model.

\subsection{Indus Valley: VIC-Derived Drought-Restoration Intervention}

The second grounded-counterfactual study revisits the Indus Valley
Civilization example used in earlier \textsc{Democritus} work, but now with an
external hydrological substrate.  The paper \emph{River drought forcing of the
Harappan metamorphosis} \citep{solanki2025harappanMetamorphosis} combines
transient climate simulations with the Variable Infiltration Capacity (VIC)
hydrological model to reason about severe river droughts during the
Harappan transition.  The local paper folder contains the PDF, paper-specific
figure data, and an upstream VIC codebase.  The figure data include
station-level discharge anomalies for the four drought events, rainfall and SPI
time series, Indus-region shapefiles, Harappan site locations, and drought
anomaly grids.

This case is epistemically different from the microplastics example.  The
folder does not contain the full transient Indus forcing grids needed to
reproduce the authors' complete basin-scale VIC experiment.  It does,
however, contain the paper's VIC-derived discharge-anomaly table and the VIC
codebase itself.  We therefore treat the figure data as the paper-grounded
evaluation substrate and the local VIC run as an executable hydrology harness
that confirms the intervention path.  This distinction is recorded in the
artifact, so the result is not misreported as a full reproduction of the
authors' simulation.

\paragraph{Language topos.}
The baseline Indus claim world model contains a compact causal chain:
transient climate forcings drive VIC hydrological reconstructions; the D3
rainfall deficit and warming reduce river flow; persistent river drought
reduces freshwater availability; reduced water availability may contribute to
population dispersal from major Harappan centers; and social and economic
pressures also shape the transformation.  This last event is important: the
atlas does not collapse the historical explanation into a monocausal climate
claim.  The hydrology layer is one chart in a larger explanation.

\begin{figure}[t]
\vspace{-0.6em}
\centering
\begin{minipage}[t]{0.48\linewidth}
\centering
\begin{tikzpicture}[
  ctx/.style={draw, rounded corners=2pt, align=center, font=\tiny, inner sep=2pt, minimum width=2.35cm, minimum height=0.50cm},
  good/.style={-{Latex[length=2mm]}, line width=0.55pt, draw=teal!70!black},
  tense/.style={-{Latex[length=2mm]}, line width=0.65pt, draw=red!70!black, dashed}
]
\node[ctx, fill=gray!10] (corpus) at (0,1.85) {Indus claim\\topos\\5 events};
\node[ctx, fill=teal!8] (vic) at (-1.7,0.75) {VIC\\hydrology\\1};
\node[ctx, fill=teal!8] (multi) at (1.7,0.75) {multi-factor\\explanation\\1};
\node[ctx, fill=red!8] (monsoon) at (-1.85,-0.45) {restored\\monsoon\\1};
\node[ctx, fill=orange!10] (water) at (0,-0.72) {freshwater\\availability\\1};
\node[ctx, fill=orange!10] (har) at (1.85,-0.45) {Harappan\\metamorphosis\\1};
\draw[good] (corpus) -- (vic);
\draw[good] (corpus) -- (multi);
\draw[tense] (corpus) -- (monsoon);
\draw[tense] (corpus) -- (water);
\draw[tense] (corpus) -- (har);
\node[font=\tiny, fill=white, inner sep=0.7pt, text=teal!70!black] at (-2.45,1.32) {$\Delta=.041$};
\node[font=\tiny, fill=white, inner sep=0.7pt, text=teal!70!black] at (2.45,1.32) {$\Delta=.057$};
\node[font=\tiny, fill=white, inner sep=0.7pt, text=red!70!black] at (-2.95,0.25) {$\Delta=.093$};
\node[font=\tiny, fill=white, inner sep=0.7pt, text=red!70!black] at (-0.55,0.18) {$\Delta=.093$};
\node[font=\tiny, fill=white, inner sep=0.7pt, text=red!70!black] at (2.95,0.25) {$\Delta=.407$};
\node[font=\tiny, align=left] at (0,-1.35) {solid: compatible; dashed: hydrology-layer revision};
\end{tikzpicture}
\end{minipage}
\hfill
\begin{minipage}[t]{0.48\linewidth}
\centering
\begin{tikzpicture}[x=1cm,y=1cm]
\draw[->, line width=0.45pt] (0,0) -- (4.8,0);
\draw[->, line width=0.45pt] (0,0) -- (0,2.25);
\foreach \x/\lab in {0.95/D3,2.55/restored,4.05/recovery} {
  \node[font=\tiny] at (\x,-0.25) {\lab};
}
\fill[blue!55] (0.60,0) rectangle (1.30,1.65);
\fill[teal!55] (2.20,0) rectangle (2.90,1.80);
\fill[red!55] (3.70,0) rectangle (4.40,0.67);
\node[font=\tiny] at (0.95,1.83) {$91.51$};
\node[font=\tiny] at (2.55,1.98) {$100.00$};
\node[font=\tiny] at (4.05,0.85) {$+8.49$};
\node[font=\tiny, align=center] at (2.35,2.42) {VIC-derived discharge index};
\node[font=\tiny, align=center] at (2.40,-0.70) {rainfall deficit \(-9.79\%\), SPI \(-0.38\)};
\end{tikzpicture}
\end{minipage}
\vspace{-0.4em}
\caption{A concrete Indus Valley artifact slice.  Left: local contexts in the
counterfactual sheaf after the drought-restoration intervention.  Right: the
paper-grounded discharge-anomaly substrate moves the D3 hydrology index from
\(91.51\) to \(100.00\), while the rainfall/SPI source data supply a
climate-side consistency check.}
\label{fig:indus-artifact-sheaf}
\vspace{-0.8em}
\end{figure}
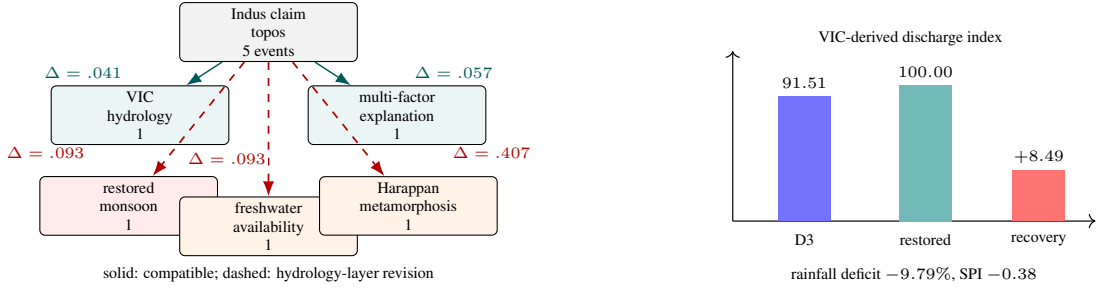

\paragraph{Grounded intervention.}
The executable question is a drought-restoration counterfactual:
\[
  \text{D3 rainfall deficit and warming}
  \longmapsto
  \text{restored precipitation and baseline temperature}.
\]
From the paper's discharge-anomaly source table, D3 has a mean discharge
anomaly of \(-8.49\%\) across \(18\) stations, with station anomalies ranging
from \(-15.37\%\) to \(-5.22\%\).  We normalize the D3 discharge state as an
index of \(91.51\), where \(100\) denotes restored baseline discharge.  The
grounded counterfactual therefore recovers \(8.49\) index points, or \(9.28\%\)
relative to the D3 drought state.  The rainfall/SPI source-data panel provides
a consistent climate-side check: over the D3 interval, mean rainfall is
approximately \(312.62\,\mathrm{mm}\) versus a Pre-Harappan reference mean of
\(346.53\,\mathrm{mm}\), a \(-9.79\%\) deficit, with mean SPI \(-0.38\).
\Cref{fig:indus-artifact-sheaf} summarizes the induced sheaf revision and the
paper-grounded hydrology index shift.

\paragraph{Back into the sheaf.}
As in the microplastics case, the numeric result is pushed back into the world
model.  The baseline observation
\begin{quote}
\footnotesize\ttfamily
d3\_rainfall\_deficit|reduces|river\_flow
\end{quote}
is replaced by the counterfactual observation
\begin{quote}
\footnotesize\ttfamily
counterfactual\_restored\_monsoon\_forcing|increases|\\
vic\_water\_availability\_proxy.
\end{quote}
Freshwater-availability and Harappan-metamorphosis events are also rewritten,
so the rebuilt model contains counterfactual contexts for restored monsoon
forcing, freshwater availability, and the weakened hydrology-only support for
drought-driven dispersal.  In the generated artifact, three causal events are
modified and the rebuilt counterfactual world model contains \(5\) events,
\(6\) local PSRs, \(6\) sheaf objects, \(5\) restriction checks, and \(5\)
gluing diagnostics.  The grounded layer reports the primary metric as a
\emph{VIC-derived discharge index}: baseline \(91.51\), counterfactual
\(100.00\), effect \(8.49\).

\subsection{Sachs: Experimental-Regime Substitution in Protein Signaling}

The third grounded case uses the canonical Sachs et al. protein-signaling
study \citep{sachs2005causalProteinSignaling}.  This domain is especially
useful for \textsc{Prometheus} because it is both a famous causal-discovery
benchmark and an experimentally perturbed biological system.  We first ran the
paper through the standard claims pipeline, producing a Sachs claims atlas
centered on T-cell receptor stimulation, phosphorylation cascades, Bayesian
network causal inference, and signaling proteins such as Raf, Mek, PLC-\(\gamma\),
PIP2, PIP3, Erk, Akt, PKA, PKC, p38, and Jnk.  We then used the local
\texttt{sachs\_with\_env.csv} panel as the external substrate.  The file
contains \(853\) single-cell observations over \(11\) markers, partitioned
into four anonymous experimental environments: \(e0\) with \(88\) rows, \(e1\)
with \(218\), \(e2\) with \(467\), and \(e3\) with \(80\).

Because the local data file preserves environment labels but not the full
paper-condition names, the intervention is stated conservatively as an
environment substitution rather than as a named inhibitor or activator.  Taking
the largest regime \(e2\) as the baseline, \textsc{Prometheus} computes marker
means on a \(\log(1+x)\) abundance scale and evaluates substitutions from \(e2\)
to each other environment.  The strongest coherent marker-index shift is
\[
  e2 \longmapsto e0,
\]
where the mean log-abundance index over the three most shifted markers
PKA/Akt/Erk changes from \(3.968\) to \(4.993\), a shift of \(+1.025\)
log-abundance units, or \(25.8\%\) relative to the baseline index.  The
canonical Sachs network overlay makes the result biologically interpretable
without claiming to re-infer the network: the largest measured co-shifts occur
on \( \mathrm{PKA}\!\to\!\mathrm{Akt}\) and
\( \mathrm{PKA}\!\to\!\mathrm{Erk}\), with co-shift scores \(1.25\) and
\(1.06\), respectively.

This measured substitution is then pushed back into the Sachs claims world
model.  Thirty extracted signaling claims mentioning the shifted marker family
are rewritten into counterfactual observations, and a data-grounded episode is
added for the measured marker shifts and canonical-edge overlay.  The rebuilt
counterfactual world model contains \(383\) causal events, \(22\) local PSRs,
\(22\) sheaf objects, \(21\) restriction checks, and \(21\) gluing diagnostics.
The new local contexts include
\texttt{counterfactual\_sachs\_environment\_shift},
\texttt{sachs\_measured\_environment\_shift}, and
\texttt{sachs\_canonical\_network\_overlay}.  Thus Sachs supplies a third
kind of grounding: not source-code execution, and not a hydrology simulation,
but measured perturbation data from a benchmark causal system.

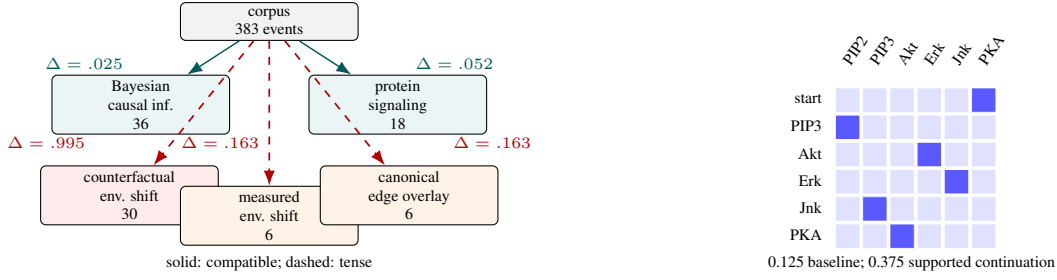
\begin{figure}[t]
\vspace{-0.6em}
\centering
\begin{minipage}[t]{0.47\linewidth}
\centering
\begin{tikzpicture}[
  ctx/.style={draw, rounded corners=2pt, align=center, font=\tiny, inner sep=2pt, minimum width=2.35cm, minimum height=0.50cm},
  good/.style={-{Latex[length=2mm]}, line width=0.55pt, draw=teal!70!black},
  tense/.style={-{Latex[length=2mm]}, line width=0.65pt, draw=red!70!black, dashed}
]
\node[ctx, fill=gray!10] (corpus) at (0,1.95) {corpus\\383 events};
\node[ctx, fill=teal!8] (bn) at (-1.7,0.85) {Bayesian\\causal inf.\\36};
\node[ctx, fill=teal!8] (sig) at (1.7,0.85) {protein\\signaling\\18};
\node[ctx, fill=red!8] (cf) at (-1.85,-0.35) {counterfactual\\env. shift\\30};
\node[ctx, fill=orange!10] (meas) at (0,-0.62) {measured\\env. shift\\6};
\node[ctx, fill=orange!10] (edge) at (1.85,-0.35) {canonical\\edge overlay\\6};
\draw[good] (corpus) -- (bn);
\draw[good] (corpus) -- (sig);
\draw[tense] (corpus) -- (cf);
\draw[tense] (corpus) -- (meas);
\draw[tense] (corpus) -- (edge);
\node[font=\tiny, fill=white, inner sep=0.7pt, text=teal!70!black] at (-2.45,1.40) {$\Delta=.025$};
\node[font=\tiny, fill=white, inner sep=0.7pt, text=teal!70!black] at (2.45,1.40) {$\Delta=.052$};
\node[font=\tiny, fill=white, inner sep=0.7pt, text=red!70!black] at (-2.95,0.35) {$\Delta=.995$};
\node[font=\tiny, fill=white, inner sep=0.7pt, text=red!70!black] at (-0.65,0.35) {$\Delta=.163$};
\node[font=\tiny, fill=white, inner sep=0.7pt, text=red!70!black] at (2.95,0.35) {$\Delta=.163$};
\node[font=\tiny, align=left] at (0,-1.25) {solid: compatible; dashed: tense};
\end{tikzpicture}
\end{minipage}
\hfill
\begin{minipage}[t]{0.49\linewidth}
\centering
\begin{tikzpicture}[x=0.36cm,y=0.36cm]
\foreach \x/\lab in {0/PIP2,1/PIP3,2/Akt,3/Erk,4/Jnk,5/PKA} {
  \node[font=\tiny, rotate=55, anchor=west] at (\x+0.2,6.55) {\lab};
}
\foreach \y/\lab in {0/start,1/PIP3,2/Akt,3/Erk,4/Jnk,5/PKA} {
  \node[font=\tiny, anchor=east] at (-0.15,5.5-\y) {\lab};
}
\foreach \y in {0,...,5} {
  \foreach \x in {0,...,5} {
    \fill[blue!12] (\x,5-\y) rectangle ++(0.9,0.9);
    \draw[white,line width=0.35pt] (\x,5-\y) rectangle ++(0.9,0.9);
  }
}
\foreach \x/\y in {5/0,0/1,3/2,4/3,1/4,2/5} {
  \fill[blue!65] (\x,5-\y) rectangle ++(0.9,0.9);
  \draw[white,line width=0.35pt] (\x,5-\y) rectangle ++(0.9,0.9);
}
\node[font=\tiny] at (2.8,-0.55) {0.125 baseline; 0.375 supported continuation};
\end{tikzpicture}
\end{minipage}
\vspace{-0.4em}
\caption{A concrete Sachs artifact slice.  Left: five local contexts from the
counterfactual Sachs sheaf, with corpus restriction gaps \(\Delta\).  Right: a
\(6\times6\) Hankel-style PSR excerpt for
\texttt{sachs\_measured\_environment\_shift}, whose rows are histories and
columns are tests for the measured \(e2\!\to\!e0\) marker shifts.}
\label{fig:sachs-artifact-sheaf-psr}
\vspace{-0.8em}
\end{figure}

For readers used to DAG-based causal discovery, Figure~\ref{fig:sachs-artifact-sheaf-psr}
should not be read as a causal graph whose nodes are biological variables and
whose arrows are causal effects.  The DAG-like objects live inside local
charts: a local chart may contain an extracted DAG, a Bayesian network, a
mechanistic simulator, a data panel, or a local PSR.  The displayed graph is a
cover/restriction diagram over those local causal worlds.  Thus
\textsc{Prometheus} does not discard DAG models; it treats them as one kind of
local causal artifact that can be glued, compared, revised, or marked
non-transportable inside a larger sheaf world model.  The aim is therefore not
to recover a single global DAG, but to build a small universe of local DAG-like
models, together with the restriction maps that show how they assemble, fail to
assemble, or require revision inside the topos world model.

\subsection{Singing Mice: MAPseq Projection-Attenuation Intervention}

The fourth grounded case moves from climate, hydrology, and cell-signaling
benchmarks to comparative neuroscience.  Isko et al. report a specific
expansion of motor cortical projections in the Alston's singing mouse
(\emph{Scotinomys teguina}) relative to the laboratory mouse
(\emph{Mus musculus}) \citep{isko2026singingMouseNature}.  The accompanying
Dryad dataset \citep{isko2026singingMouseDryad} contains MAPseq matrices for
\(12\) animals: \(5\) lab mice (MMus) and \(7\) singing mice (STeg).  Each
matrix row is a barcoded motor-cortical neuron and each column is a target
brain region, with raw counts, binarized counts, and spike-in-normalized count
matrices.  This makes the paper a useful test of a different
grounded-counterfactual pattern: the substrate is not a simulator, but a
single-neuron projection atlas tied to a behavioral-evolution claim.

The baseline \textsc{Prometheus} world model contains a compact causal bridge:
MAPseq barcodes measure target-region projections; STeg singing mice differ
from MMus lab mice in motor-cortical projection structure; STeg motor-cortex
pathways expand toward the auditory region (AUD) and periaqueductal gray
(PAG); and the AUD/PAG-biased expansion supports the claim that motor-cortex
pathway expansion broadens the vocal repertoire.  The final bridge is marked
as evidential rather than as a direct behavioral simulation: the local
executable substrate is a projection matrix, not a full acoustic or
vocal-motor dynamics model.

The intervention asks what happens to the claim support if the singing-mouse
AUD/PAG projection layer is attenuated to the lab-mouse species mean:
\[
  \text{STeg AUD/PAG projection support}
  \longmapsto
  \text{MMus-like AUD/PAG projection support}.
\]
Using the binarized MAPseq matrices, \textsc{Prometheus} computes per-animal
species means for the fraction of barcoded neurons with a positive projection
to each target.  The AUD fraction changes from \(0.042\) in MMus to \(0.131\)
in STeg, a \(3.10\times\) increase.  The PAG fraction changes from \(0.017\)
to \(0.079\), a \(4.61\times\) increase.  Summing the two focus targets gives
an AUD+PAG support index of
\[
  0.210 \quad \text{for STeg}
  \qquad \text{versus} \qquad
  0.059 \quad \text{for the MMus-like counterfactual}.
\]
Thus the species-level projection-attenuation counterfactual reduces the
dataset-backed AUD+PAG support by \(0.151\), or approximately \(71.7\%\).

This measured result is then pushed back into the causal observation stream.
The baseline observations
\begin{quote}
\footnotesize\ttfamily
steg\_motor\_cortex\_pathway|expands\_to|auditory\_region
\end{quote}
and
\begin{quote}
\footnotesize\ttfamily
steg\_motor\_cortex\_pathway|expands\_to|periaqueductal\_gray
\end{quote}
are replaced by counterfactual observations in which STeg AUD and PAG
projection support attenuates to the MMus species mean.  The vocal-repertoire
bridge is also rewritten as
\begin{quote}
\footnotesize\ttfamily
counterfactual\_projection\_attenuation|weakens|\\
vocal\_repertoire\_claim\_support.
\end{quote}
The rebuilt counterfactual world model contains \(5\) causal events, \(6\)
local PSRs, \(6\) sheaf objects, \(5\) restriction checks, and \(5\) gluing
diagnostics.  The new local contexts include
\texttt{counterfactual\_auditory\_projection\_expansion},
\texttt{counterfactual\_pag\_vocal\_motor\_projection}, and
\texttt{counterfactual\_vocal\_repertoire\_claim\_bridge}.  This case shows
that grounded counterfactuals need not be limited to papers with source code:
when a paper ships a structured scientific dataset, \textsc{Prometheus} can
turn a verbal mechanism claim into a measured, dataset-backed sheaf revision.

Together, the microplastics, Indus, Sachs, and singing-mouse studies make the
grounded-counterfactual contract much stronger.  One example intervenes on
optical forcing in source tables and gridded climate-forcing data.  A second
intervenes on a palaeohydrological drought mechanism using VIC-derived
discharge data and a VIC executable harness.  A third substitutes experimental
regimes in a single-cell causal-discovery benchmark.  A fourth attenuates a
species-specific motor-projection expansion in a comparative-neuroscience
dataset.  In all four cases, the measured result is not a visual
annotation layered on top of the atlas.  It changes the observations from
which the local PSRs, sheaf objects, restrictions, and gluing diagnostics are
rebuilt.

\section{Evaluation}

\textsc{Prometheus} should be evaluated as a causal research instrument.
Extraction accuracy matters, but it is only one part of the story.  The case
studies above require metrics that cover both the text-to-atlas pipeline and
the grounded-counterfactual loop.  We propose the following evaluation axes.

\begin{table}[t]
\centering
\begin{tabular}{p{0.25\linewidth}p{0.62\linewidth}}
\toprule
Axis & Question \\
\midrule
Claim quality & Are extracted cause, effect, modifier, polarity, and provenance correct? \\
Coverage & Does the atlas recover the major causal regions and not only the most frequent claims? \\
Support aggregation & Are repeated claims combined without erasing important context? \\
Drift visibility & Does the atlas expose changes across time, retrieval runs, or document strata? \\
Gluing usefulness & Do reported tensions correspond to real contradictions, regime boundaries, or missing evidence? \\
Provenance quality & Can users trace atlas claims back to source evidence quickly and reliably? \\
Grounding quality & When data, code, or scientific models exist, is the executed intervention faithful to the source artifact and are its limits recorded? \\
Expert navigation time & Do domain experts answer deep corpus questions faster or with fewer missed caveats? \\
Rerun consistency & Are stable claims stable across repeated retrieval and extraction runs, while true new evidence appears as change? \\
\bottomrule
\end{tabular}
\caption{\textsc{Prometheus} evaluation should measure navigational and epistemic value,
not only benchmark extraction accuracy.}
\label{tab:evaluation}
\end{table}

Some axes can be scored automatically.  Claim quality can use annotated causal
extraction sets.  Rerun consistency can compare atlas topology and claim tables
across seeds.  Support aggregation can be tested against synthetic corpora with
known repeated local claims.  Grounding quality can be checked against
source-data hashes, executable provenance, intervention parameters, and whether
the regenerated sheaf records the difference between full reproduction and
partial figure-data grounding.  Other axes require expert studies.  For
example, domain experts can be asked to answer multi-hop literature questions
with and without the atlas, measuring time, evidence recall, missed regime
caveats, and whether the system correctly exposes the limits of current
knowledge.

\section{Limitations}

\textsc{Prometheus} inherits the weaknesses of its sources and extractors.  Retrieval
drift can change the corpus before modeling begins.  LLM extraction can be
redundant, overconfident, or sensitive to prompt wording.  Claim
canonicalization remains difficult: two passages may express the same relation
with different variables, or different relations with deceptively similar
language.  Source quality matters; a gluing procedure cannot make weak evidence
strong.  Local intervention probes are model-internal tests unless paired with
identification assumptions or external validation.  The microplastics and
Indus counterfactuals in \cref{sec:grounded-counterfactuals} illustrate the
stronger case where external source data, figure data, simulation outputs, or
code exist, but many papers will not provide a runnable or source-table
substrate suitable for this kind of grounding.  Even when source artifacts
exist, the grounding can be partial: the Indus folder contains paper-specific
VIC-derived figure data and a VIC codebase, but not the complete transient
forcing grids needed to reproduce the original basin-scale simulation.

Cost is also a practical limitation.  Corpus-scale extraction, provenance
tracking, and repeated reruns can be expensive.  \textsc{Prometheus} therefore needs
caching, incremental updates, model routing, and compact artifact schemas.

Finally, \textsc{Prometheus} is not meant to remove human steering.  It is designed to
make steering more informed: users should be able to exclude regions, split
contexts, revise canonicalization, inspect provenance, and decide which gluing
tensions deserve follow-up.

\section{Research Roadmap}

The next stage is interactive refinement.  Users should be able to enter a
region, exclude unreliable sources, split a context, merge synonymous claims,
and ask for a persistent-state comparison against an earlier atlas.  Active
corpus steering should let the system propose retrievals that would reduce a
specific gluing tension.  Stronger claim canonicalization should combine
embedding, symbolic normalization, ontology hints, and human corrections.

On the modeling side, \textsc{Prometheus} should move from deterministic overlap
diagnostics toward learned restriction maps, neural or kernel PSR estimators,
uncertainty-aware gluing, and explicit sheafification procedures.  On the
interface side, the Claims Atlas should become a live research surface: causal
spines, local regions, provenance, drift, and repair suggestions should be
visible as first-class objects.  The grounded-counterfactual layer adds another
roadmap item: when papers include code, tables, simulations, figure-source
data, or agent-executable action hooks, \textsc{Prometheus} should learn to
discover executable intervention hooks, run them, distinguish full
reproductions from partial figure-data groundings, and rebuild the affected
sheaf charts automatically.

\section{Future Directions: Substrate-Seeking Topos Construction}

The grounded case studies also suggest a broader goal for
\textsc{Prometheus}~v2 and beyond.  Scientific inquiry rarely begins with a
single document.  It begins with a question: Why did the Indus Valley
Civilization transform?  What is the climatic effect of airborne
microplastics?  Which mechanisms explain a drug's benefit or risk?  Scientists
then marshal heterogeneous evidence---papers, proxy records, experiments,
source tables, equations, simulation outputs, model code, figures, and
caveats---and attempt to assemble a coherent explanation.  In categorical
terms, a scientific paper can be read as a compressed topos world model: a
cover of local evidentiary charts together with an argument for how some of
those charts glue into a global explanatory section.

The charts need not glue perfectly.  A hydrological model may support a drought
hypothesis while archaeological timing leaves room for social mechanisms.  A
radiative-transfer calculation may support an optical-forcing claim while
particle aging, spatial distribution, or measurement assumptions remain local
sources of uncertainty.  Experiments may validate one mechanism and leave
another underdetermined.  These non-gluing regions are not failures of the
research process.  They are often the most valuable output: they mark the
limits of current knowledge and identify where new data, new measurements, or
new interventions are needed.

This is where \textsc{Prometheus} should move beyond AI-enabled search engines.
Search and retrieval systems can find relevant documents, synthesize textual
answers, and summarize consensus.  Some can cite sources and sketch plausible
conclusions.  But they generally do not construct an explicit world model whose
local sections can be compared, whose failures to glue are visible, and whose
claims can be revised by running a counterfactual against data or scientific
models.  They answer the question, but they do not usually expose the geometry
of the evidence: which local worlds support the answer, which contradict it,
which are merely compatible, and which remain unknown.

Substrate-seeking retrieval is therefore a central future direction.  Starting
from a causal query, \textsc{Prometheus} should not only retrieve text.  It
should search for the surrounding research substrate: supplementary tables,
figure data, notebooks, repositories, simulation inputs, package versions,
model checkpoints, data DOIs, experimental protocols, and agent-executable
tools.  Each artifact becomes a candidate chart in the topos world model.  Text
provides causal discourse and hypotheses; data and figures provide measured
sections; code and scientific models provide executable transition rules; and
the sheaf layer records how these local worlds agree, disagree, or fail to
transport across regimes.

In this view, the long-term promise of \textsc{Prometheus} is not merely
better summarization.  It is the automation of a broader scientific practice:
constructing, testing, and revising causal world models from heterogeneous
evidence.  A mature system should be able to say not only ``here is the best
answer supported by the current corpus,'' but also ``here are the local worlds
that support it, here are the worlds that obstruct it, here is the
counterfactual we can actually run, and here is the boundary beyond which the
present evidence does not justify a conclusion.''

\section{Conclusion}

\textsc{Prometheus} reframes causal research as the construction of a Topos
World Model: a sheaf-like atlas of local causal predictive states over a
heterogeneous research substrate.  The point is not to produce one more flat
summary.  It is to preserve locality, support, drift, contradiction,
provenance, and epistemic limits while giving researchers a navigable causal
structure.  Large language models can extract local causal claims;
\textsc{Prometheus} asks how those claims live together, where they fail to
glue, and how those failures can guide deeper research.  The grounded
microplastics, Indus, Sachs, and singing-mouse counterfactuals show the next
step: when a research substrate contains both language and external data or
scientific models, the atlas can move from internal counterfactual probes to
measured interventions and then rebuild the local world around the result.

\section*{Code Availability}

The predecessor \textsc{Democritus} codebase is publicly available as the
\texttt{Democritus\_OpenAI} repository
\citep{mahadevanDemocritusOpenAI}.  \textsc{Prometheus} currently builds on
this released causal-extraction lineage but adds an actively developing product
layer for topos world-model construction, Claims Atlas navigation, persistent
state, and grounded counterfactual execution.  We therefore do not release the
\textsc{Prometheus} code with this manuscript.  A public code release is planned
once the system has matured into a stable, documented research product.

\appendix

\section{System Genealogy and GUI Modes}
\label{app:system-genealogy}

\textsc{Prometheus} inherits part of its interface genealogy from our earlier
CLIFF chatbot and local research interface
\citep{mahadevanCLIFFCatAgi}.  CLIFF began as a Categories-for-AGI companion
system for interactive retrieval, teaching, and research workflows.  The
\textsc{Prometheus} GUI reuses several lessons from that system: a
natural-language query box, long-running local sessions, background execution,
artifact dashboards, route-specific reports, and persistent run directories.
The conceptual boundary is different.  CLIFF remains oriented toward the
Categories for AGI book, course material, and general retrieval-conditioned
chatbot workflows, whereas \textsc{Prometheus} is reserved for causal research
artifacts, local PSR construction, gluing diagnostics, persistent world state,
and Claims Atlas navigation.

The GUI is designed to accept broad natural-language research requests and route
them to specialized workflows.  In the current implementation, route families
include literature and paper-corpus synthesis, Democritus-style causal-claim
analysis, SEC and company-filing workflows, product-feedback world models,
targeted-sentiment review benchmarks, Rock--Paper--Scissors and network-economy
agent traces, and small Topos/OOM experiments.  A route may emit several
artifacts: a human-readable report, a technical dashboard, a JSON world-model
bundle, a persistent-state file, and auxiliary provenance or Claims Atlas HTML.

The GUI exposes three execution modes.  \emph{Quick} mode runs the most compact
version of a workflow and is useful for smoke tests or shallow artifact
inspection.  \emph{Interactive} mode keeps the local session open while
background runs complete, letting a researcher launch follow-up queries and
inspect completed artifacts from the session list.  \emph{Deep} mode allocates
more work to acquisition, extraction, synthesis, and report generation, and is
the intended setting for the case-study style runs described in this paper.
The GUI also exposes an analysis-mode choice: \emph{standard} runs the routed
workflow in its ordinary reporting mode, while \emph{Topos World Model} attaches
the \textsc{Prometheus} layer when supported, producing local PSRs, restrictions,
gluing diagnostics, and persistent-state artifacts.

Several additional controls specialize particular routes rather than changing
the overall framework.  Democritus-style claim analysis can run with full,
lightweight, or mixture-of-experts manifold modes, optional dry-run behavior,
and optional deep-dive report generation.  Filing workflows can use dry-run
paths for debugging.  Product-feedback and persistent-state workflows can take a
parent state or state query, allowing a follow-up run to compare against an
earlier world model.  These options are engineering controls, not separate
theoretical models; they let users trade runtime, cost, and depth while keeping
the same artifact contract.


\begin{thebibliography}{25}
\providecommand{\natexlab}[1]{#1}
\providecommand{\url}[1]{\texttt{#1}}
\expandafter\ifx\csname urlstyle\endcsname\relax
  \providecommand{\doi}[1]{doi: #1}\else
  \providecommand{\doi}{doi: \begingroup \urlstyle{rm}\Url}\fi

\bibitem[Abramsky and Brandenburger(2011)]{abramsky2011sheaf}
Samson Abramsky and Adam Brandenburger.
\newblock The sheaf-theoretic structure of non-locality and contextuality.
\newblock \emph{New Journal of Physics}, 13\penalty0 (11):\penalty0 113036,
  2011.

\bibitem[Girju(2003)]{girju2003automatic}
Roxana Girju.
\newblock Automatic detection of causal relations for question answering.
\newblock In \emph{Proceedings of the ACL Workshop on Multilingual
  Summarization and Question Answering}, 2003.

\bibitem[Hassanzadeh et~al.(2020)Hassanzadeh, Bhattacharjya, Feblowitz,
  Perrone, Sohrabi, Srinivas, and Katz]{hassanzadeh2020causalKB}
Oktie Hassanzadeh, Debarun Bhattacharjya, Mark Feblowitz, Michael Perrone,
  Shirin Sohrabi, Kavitha Srinivas, and Michael Katz.
\newblock Causal knowledge extraction through large-scale text mining.
\newblock In \emph{Proceedings of the AAAI Conference on Artificial
  Intelligence}, volume~34, pages 13520--13527, 2020.

\bibitem[He et~al.(2023)He, Guan, and Chen]{he2022eventCausalitySurvey}
Xiaomei He, Yi~Guan, and Min Chen.
\newblock A survey of event causality identification: Taxonomy, resources, and
  techniques.
\newblock \emph{ACM Computing Surveys}, 55\penalty0 (14s):\penalty0 1--35,
  2023.
\newblock \doi{10.1145/3582128}.

\bibitem[Hendrickx et~al.(2010)Hendrickx, Kim, Kozareva, Nakov,
  {\'O}~S{\'e}aghdha, Pad{\'o}, Pennacchiotti, Romano, and
  Szpakowicz]{hendrickx2010semeval}
Iris Hendrickx, Su~Nam Kim, Zornitsa Kozareva, Preslav Nakov, Diarmuid
  {\'O}~S{\'e}aghdha, Sebastian Pad{\'o}, Marco Pennacchiotti, Lorenza Romano,
  and Stan Szpakowicz.
\newblock {SemEval-2010} task 8: Multi-way classification of semantic relations
  between pairs of nominals.
\newblock In \emph{Proceedings of the 5th International Workshop on Semantic
  Evaluation}, pages 33--38, 2010.

\bibitem[Isko et~al.(2026{\natexlab{a}})Isko, Harpole, Zheng, Zhan, Davis,
  Zador, and Banerjee]{isko2026singingMouseDryad}
Emily Isko, Clifford Harpole, Xiaoyue~Mike Zheng, Huiqing Zhan, Martin Davis,
  Anthony Zador, and Arkarup Banerjee.
\newblock Data from: Specific expansion of motor cortical projections in a
  singing mouse.
\newblock Dryad dataset, 2026{\natexlab{a}}.

\bibitem[Isko et~al.(2026{\natexlab{b}})Isko, Harpole, Zheng, Zhan, Davis,
  Zador, and Banerjee]{isko2026singingMouseNature}
Emily~C. Isko, Clifford~E. Harpole, Xiaoyue~Mike Zheng, Huiqing Zhan, Martin~B.
  Davis, Anthony~M. Zador, and Arkarup Banerjee.
\newblock Specific expansion of motor cortical projections in a singing mouse.
\newblock \emph{Nature}, 2026{\natexlab{b}}.
\newblock \doi{10.1038/s41586-026-10458-y}.
\newblock Published May 6, 2026.

\bibitem[Jin et~al.(2021)Jin, Sch{\"o}lkopf, Spirtes, and
  Zhang]{jin2021causalnlpSurvey}
Zhijing Jin, Bernhard Sch{\"o}lkopf, Peter Spirtes, and Kun Zhang.
\newblock Causal inference and natural language processing: A survey.
\newblock \emph{arXiv preprint arXiv:2012.14366}, 2021.

\bibitem[K{\i}c{\i}man et~al.(2024)K{\i}c{\i}man, Ness, Sharma, and
  Tan]{kiciman2024causalLLM}
Emre K{\i}c{\i}man, Robert~Osazuwa Ness, Amit Sharma, and Chenhao Tan.
\newblock Causal reasoning and large language models: Opening a new frontier
  for causality.
\newblock \emph{Transactions on Machine Learning Research}, 2024.
\newblock URL \url{https://openreview.net/forum?id=6z4djmZK3c}.
\newblock Preprint arXiv:2305.00050.

\bibitem[Le et~al.(2024)Le, Xia, and Chen]{le2024multiagentCausalDiscovery}
Hao~Duong Le, Xin Xia, and Zhang Chen.
\newblock Multi-agent causal discovery using large language models.
\newblock \emph{arXiv preprint arXiv:2407.15073}, 2024.

\bibitem[Lewis et~al.(2020)Lewis, Perez, Piktus, Petroni, Karpukhin, Goyal,
  Kuttler, Lewis, Yih, Rocktaschel, Riedel, and Kiela]{lewis2020retrieval}
Patrick Lewis, Ethan Perez, Aleksandra Piktus, Fabio Petroni, Vladimir
  Karpukhin, Naman Goyal, Heinrich Kuttler, Mike Lewis, Wen-tau Yih, Tim
  Rocktaschel, Sebastian Riedel, and Douwe Kiela.
\newblock Retrieval-augmented generation for knowledge-intensive nlp tasks.
\newblock In \emph{Advances in Neural Information Processing Systems}, 2020.

\bibitem[Littman et~al.(2001)Littman, Sutton, and Singh]{littman2001predictive}
Michael~L. Littman, Richard~S. Sutton, and Satinder Singh.
\newblock Predictive representations of state.
\newblock In \emph{Advances in Neural Information Processing Systems}, 2001.

\bibitem[Liu et~al.(2026)]{liu2026atmosphericMicroplastics}
Yu~Liu et~al.
\newblock Atmospheric warming contributions from airborne microplastics and
  nanoplastics.
\newblock \emph{Nature Climate Change}, 2026.
\newblock \doi{10.1038/s41558-026-02620-1}.
\newblock Source data DOI: 10.5281/zenodo.19042838.

\bibitem[Mac~Lane and Moerdijk(1992)]{maclane1992sheaves}
Saunders Mac~Lane and Ieke Moerdijk.
\newblock \emph{Sheaves in Geometry and Logic: A First Introduction to Topos
  Theory}.
\newblock Springer, 1992.

\bibitem[Mahadevan(2025{\natexlab{a}})]{mahadevan2025largecausalmodelslarge}
Sridhar Mahadevan.
\newblock Large causal models from large language models, 2025{\natexlab{a}}.
\newblock URL \url{https://arxiv.org/abs/2512.07796}.

\bibitem[Mahadevan(2025{\natexlab{b}})]{mahadevanCLIFFCatAgi}
Sridhar Mahadevan.
\newblock {CLIFF\_CatAgi}: Categories for {AGI} local research interface.
\newblock GitHub repository, 2025{\natexlab{b}}.
\newblock URL \url{https://github.com/sridharmahadevan/CLIFF_CatAgi}.

\bibitem[Mahadevan(2025{\natexlab{c}})]{mahadevanCatAGIBook}
Sridhar Mahadevan.
\newblock Categories for {AGI}.
\newblock Book manuscript, 2025{\natexlab{c}}.
\newblock URL \url{https://people.cs.umass.edu/~mahadeva/papers/catagi.pdf}.

\bibitem[Mahadevan(2025{\natexlab{d}})]{mahadevanDemocritusOpenAI}
Sridhar Mahadevan.
\newblock {Democritus\_OpenAI}: Whygraphs from large language models.
\newblock GitHub repository, 2025{\natexlab{d}}.
\newblock URL \url{https://github.com/sridharmahadevan/Democritus_OpenAI}.

\bibitem[Pearl(2009)]{pearl2009causality}
Judea Pearl.
\newblock \emph{Causality: Models, Reasoning, and Inference}.
\newblock Cambridge University Press, 2 edition, 2009.

\bibitem[Radinsky et~al.(2012)Radinsky, Davidovich, and
  Markovitch]{radinsky2012learningCausality}
Kira Radinsky, Sagie Davidovich, and Shaul Markovitch.
\newblock Learning causality for news events prediction.
\newblock In \emph{Proceedings of the 21st International Conference on World
  Wide Web}, pages 909--918, 2012.
\newblock \doi{10.1145/2187836.2187958}.

\bibitem[Sachs et~al.(2005)Sachs, Perez, Pe'er, Lauffenburger, and
  Nolan]{sachs2005causalProteinSignaling}
Karen Sachs, Omar Perez, Dana Pe'er, Douglas~A. Lauffenburger, and Garry~P.
  Nolan.
\newblock Causal protein-signaling networks derived from multiparameter
  single-cell data.
\newblock \emph{Science}, 308\penalty0 (5721):\penalty0 523--529, 2005.
\newblock \doi{10.1126/science.1105809}.

\bibitem[Singh et~al.(2004)Singh, James, and Rudary]{singh2004predictive}
Satinder Singh, Michael~R. James, and Matthew~R. Rudary.
\newblock Predictive state representations: A new theory for modeling dynamical
  systems.
\newblock \emph{Proceedings of the 20th Conference on Uncertainty in Artificial
  Intelligence}, 2004.

\bibitem[Solanki et~al.(2025)Solanki, Jain, Thirumalai, Rajagopalan, and
  Mishra]{solanki2025harappanMetamorphosis}
Hiren Solanki, Vikrant Jain, Kaustubh Thirumalai, Balaji Rajagopalan, and Vimal
  Mishra.
\newblock River drought forcing of the harappan metamorphosis.
\newblock \emph{Communications Earth \& Environment}, 6:\penalty0 926, 2025.
\newblock \doi{10.1038/s43247-025-02901-1}.

\bibitem[Yamada et~al.(2025)Yamada, Lange, Lu, Hu, Lu, Foerster, Clune, and
  Ha]{yamada2025aiScientistV2}
Yutaro Yamada, Robert~Tjarko Lange, Cong Lu, Shengran Hu, Chris Lu, Jakob
  Foerster, Jeff Clune, and David Ha.
\newblock The {AI} scientist-v2: Workshop-level automated scientific discovery
  via agentic tree search.
\newblock \emph{arXiv preprint arXiv:2504.08066}, 2025.
\newblock \doi{10.48550/arXiv.2504.08066}.

\bibitem[Yang et~al.(2022)Yang, Han, and Poon]{yang2022causalSurvey}
Jie Yang, Soyeon~Caren Han, and Josiah Poon.
\newblock A survey on extraction of causal relations from natural language
  text.
\newblock \emph{Knowledge and Information Systems}, 64\penalty0 (5):\penalty0
  1161--1186, 2022.
\newblock \doi{10.1007/s10115-022-01665-w}.

\end{thebibliography}
\end{document}